\documentclass[11pt,a4paper,logo,copyright]{amazon}

\usepackage[authoryear,sort&compress,round]{natbib}
\usepackage{cleveref}
\usepackage[textsize=small]{todonotes}
\usepackage[dvipsnames]{xcolor}
\usepackage{graphicx}
\usepackage{booktabs}
\usepackage{tabularx}
\usepackage{longtable}
\usepackage{makecell}
\usepackage{multirow}
\usepackage{array}
\usepackage{siunitx}
\usepackage{colortbl}
\usepackage{arydshln}
\usepackage{adjustbox}
\usepackage{wrapfig}
\usepackage{float}
\usepackage{afterpage}
\usepackage{listings}
\usepackage{xspace}
\usepackage{caption}
\usepackage{etoolbox}
\usepackage{dirtree}
\usepackage[edges]{forest}

\AtBeginEnvironment{tabular}{\footnotesize}
\AtBeginEnvironment{tabular*}{\footnotesize}

\usepackage{subcaption}

\usepackage{amsmath,amssymb}
\usepackage{bbm}
\usepackage{pifont} 

\usepackage{algorithm}
\usepackage{algpseudocode}

\usepackage[toc,page,header]{appendix}
\usepackage{minitoc}
\usepackage{placeins} 
\usepackage{titletoc}

\usepackage{enumitem}
\usepackage{parskip}  

\usepackage[most]{tcolorbox}

\usepackage{fancyvrb}
\usepackage{xurl}
\usepackage{seqsplit}  

\usepackage{hyphenat}

\let\oldunderscore\_
\renewcommand{\_}{\oldunderscore\allowbreak}

\DeclareRobustCommand{\texttt}[1]{\protect\textttbreakable{#1}}
\newcommand{\textttbreakable}[1]{{\ttfamily\seqsplit{#1}}}

\usepackage{tikz}
\usetikzlibrary{patterns,calc}

\usepackage{pgfplots}
\pgfplotsset{compat=1.18}
\usepgfplotslibrary{groupplots}
\usetikzlibrary{pgfplots.groupplots}

\definecolor{sigterm}{RGB}{0,102,204}
\definecolor{critterm}{RGB}{204,0,0}

\definecolor{commentcolor}{rgb}{0.25,0.5,0.5}
\definecolor{monokaipurple}{RGB}{145, 94, 174}
\definecolor{monokaired}{RGB}{204, 120, 50}
\colorlet{darkgreen}{green!65!black}
\colorlet{darkblue}{blue!75!black}
\colorlet{darkred}{red!80!black}
\definecolor{lightgreen}{HTML}{39b54a}
\definecolor{mypurple}{HTML}{412F8A}
\definecolor{myorange}{HTML}{fc8e62}
\definecolor{mygreen}{RGB}{57, 172, 57}
\definecolor{mytred}{RGB}{200, 10, 10}
\definecolor{mygray}{RGB}{100, 100, 100}
\definecolor{mygold}{rgb}{225, 215, 200}
\definecolor{navyblue}{RGB}{40, 66, 200}
\definecolor{lightblue}{RGB}{140, 166, 240}
\definecolor{lightgray}{gray}{0.9}
\definecolor{orange}{RGB}{255,127,80}
\definecolor{pink}{RGB}{219,112,147}
\definecolor{baselinecolor}{gray}{.95}

\newcommand{\cmark}{\textcolor{darkgreen}{\ding{51}}}
\newcommand{\xmark}{\textcolor{darkred}{\ding{55}}}

\newcommand{\vs}{\textit{vs.}\xspace}
\newcommand{\eg}{\textit{e.g.}\xspace}
\newcommand{\ie}{\textit{i.e.}\xspace}

\newcommand{\graycmidrule}[1]{%
  \arrayrulecolor{gray}\cmidrule(lr){#1}\arrayrulecolor{black}%
}
\newcommand{\graycmidruletwo}[2]{%
  \arrayrulecolor{gray}\cmidrule(lr){#1}\cmidrule(lr){#2}\arrayrulecolor{black}%
}
\newcommand{\graycmidrulethree}[3]{%
  \arrayrulecolor{gray}\cmidrule(lr){#1}\cmidrule(lr){#2}\cmidrule(lr){#3}\arrayrulecolor{black}%
}
\newcommand{\graycmidrulefour}[4]{%
  \arrayrulecolor{gray}\cmidrule(lr){#1}\cmidrule(lr){#2}\cmidrule(lr){#3}\cmidrule(lr){#4}\arrayrulecolor{black}%
}

\newlist{myenum}{enumerate}{1}
\setlist[myenum]{leftmargin=2em,itemsep=0pt,topsep=0pt,partopsep=0pt}

\title{An Empirical Study of Automating Agent Evaluation}

\correspondingauthor{Placeholder}
\reportnumber{} 
\codeurl{https://github.com/awslabs/Agent-EvalKit}

\author[$\ast$]{Kang Zhou}
\author[$\ast$]{Sangmin Woo}
\author[$\dagger$]{Haibo Ding}
\author{Kiran Ramnath}
\author{Subramanian Chidambaram}
\author{Aosong Feng}
\author{Vinayak Arannil}
\author{Muhyun Kim}
\author{Ishan Singh}
\author{Darren Wang}
\author{Zhichao Xu}
\author{Megha Gandhi}
\author{Nirmal Prabhu}
\author{Soumya Smruti Mishra}
\author{Smeet Dhakecha}
\author{Vivek Singh}
\author{Gouri Pandeshwar}
\author{Lin Lee Cheong}

\affil{AWS AI Labs}

\begin{abstract}
Agent evaluation requires assessing complex multi-step behaviors involving tool use and intermediate reasoning, making it costly and expertise-intensive.
A natural question arises: 
\textit{can frontier coding assistants reliably automate this evaluation process?} 
Our study shows that simply prompting coding assistants is insufficient for this task. 
Without domain-specific evaluation knowledge, frontier coding assistants achieve only a 30\% execution success rate and produce over-engineered evaluations averaging 12+ metrics per agent, indicating that strong coding ability does not automatically translate to reliable agent evaluation.
We introduce \textsc{EvalAgent}, an AI assistant that automates the end-to-end agent evaluation pipeline. 
\textsc{EvalAgent} encodes evaluation domain expertise as \emph{evaluation skills} (procedural instructions, reusable code and templates, and dynamically retrieved API documentation) that compose into a trace-based pipeline producing complete evaluation artifacts including metrics, executable code, and reports.
To systematically assess generated evaluations, we introduce a meta-evaluation framework alongside \textsc{AgentEvalBench}, a benchmark comprising 20 agents, each paired with evaluation requirements and test scenarios. We further propose the Eval@1 metric to measure whether generated evaluation code both executes and yields meaningful results on the first run. Our experiments show that \textsc{EvalAgent} produces focused evaluations, improving Eval@1 from 17.5\% to 65\%, and achieving 79.5\% human expert preference over baseline approaches. 
Further ablation studies show that evaluation skills are critical for handling complex evaluation: removing them causes Eval@1 to drop significantly from 65\% to 30\%.

\end{abstract}

\begin{document}

\maketitle

\renewcommand{\thefootnote}{\fnsymbol{footnote}}
\footnotetext[1]{Equal contribution}
\footnotetext[2]{Corresponding author: \texttt{hbding@amazon.com}}

\newenvironment{Itemize}{%
  \begin{itemize}[leftmargin=*,itemsep=0pt,topsep=0pt,partopsep=0pt,parsep=0pt]%
}{%
  \end{itemize}%
}

\setlength{\leftmargini}{9pt}

\section{Introduction}
Evaluation is the bottleneck in agent deployment. Autonomous AI agents now operate in high-stakes applications~\citep{xi2023rise,wang2023survey}, such as code generation~\citep{wang2024openhands}, scientific discovery~\citep{zhang2025deep}, and financial analysis~\citep{han2024enhancing}. Yet the tools to assess their quality lag far behind the tools to build them. The core difficulty is that agent evaluation differs fundamentally from LLM evaluation. While LLM evaluation focuses on output quality for single-turn or multi-turn conversations~\citep{zheng2023judging,liu2023geval}, agent evaluation must assess \textit{execution traces} containing sequences of reasoning steps, tool invocations, error recovery actions, and state transitions that unfold over time~\citep{kim2025beyond,he2025traject}. Only evaluating the final output is insufficient because (1) a correct final answer may mask flawed reasoning~\citep{lightman2023lets} and (2) a failed output may not reflect robust error handling. These behavioral nuances are invisible to output-only methods and are precisely what production deployments need to assess.

Agent evaluation also demands significant domain expertise and imposes practical overhead. Practitioners must carefully design evaluation criteria tailored to diverse agent behaviors, instrument agent code to capture relevant signals, and manually inspect execution traces to diagnose failures~\citep{yehudai2025survey,mohammadi2025evaluation}. A common workaround is to embed evaluation logic directly within agent code~\citep{deepeval2024,langfuse}. However, this tightly couples evaluation with implementation, causing evaluation behavior to diverge between development and production environments. 

These challenges point to a clear need: automating the agent evaluation process itself. 
Prior work on LLM-as-a-judge~\citep{zheng2023judging,liu2023geval} and agentic judge~\citep{zhuge2024agent,kim2025beyond,he2025traject,chen2025multi} uses models to directly produce verdicts against pre-defined criteria. But judgment presupposes that criteria already exist. We identify a harder problem: \emph{evaluation generation}, given an agent's source code and execution traces, automatically generating the evaluation criteria, executable assessment code, and actionable reports. This requires not just assessing quality, but designing \textit{what} to assess and \textit{how}, a challenge that existing approaches leave unaddressed.

Can frontier coding assistants (\eg, Claude Code~\citep{anthropic2025claude45sonnet}) automate this process? Our study shows that they cannot do so reliably without intervention, exhibiting three systematic failure modes. First, \emph{metric proliferation}: without evaluation-specific guidance, they generate 12+ scattered metrics per agent, favoring operational measurements like latency percentiles and token counts over meaningful assessments of task success. Second, \emph{code over-engineering}: generated evaluation code is 2--3$\times$ longer than necessary, fragmented across 5--7 files with premature abstractions that undermine maintainability. Third, \emph{plan-code drift}: they generate evaluation plans with well-reasoned semantic rubrics but implement shallow keyword-counting heuristics instead. These failures reflect a fundamental mismatch between general code generation and the specialized reasoning that evaluation design demands.

We present \textbf{\textsc{EvalAgent}} (\S\ref{sec:method}), a system that transforms agent source code and user requirements into executable evaluation artifacts. \textsc{EvalAgent} encodes evaluation domain expertise as \emph{evaluation skills}---self-contained packages of procedural instructions, reusable code and templates, and dynamically retrieved API documentation~\citep{zhang2025agentskills}---that compose into a trace-based pipeline: planning evaluation criteria, generating test scenarios, instrumenting the agent for trace collection, producing executable assessment code, and generating actionable reports. Each pipeline stage loads only the relevant skills on demand, channeling generation toward focused, implementable assessments rather than the scattered metrics and shallow heuristics produced by unconstrained approaches.

A key challenge remains: \textit{how do we know that \textsc{EvalAgent} produces good evaluations?} To assess evaluation quality without ground truth, we introduce a \textbf{meta-evaluation framework} (\S\ref{sec:metaeval}) using pairwise comparison across five dimensions (requirement fulfillment, metric relevance, code quality, plan quality, and plan-code alignment) alongside \textbf{AgentEvalBench}, a benchmark comprising 20 diverse agents spanning 9 frameworks and 14 application domains, each paired with 2 user requirements and 5 test scenarios. Expert annotators achieve Fleiss' $\kappa = 0.923$ and prefer \textsc{EvalAgent} in 79.5\% (10.5\% ties) of comparisons. Across all conditions, \textsc{EvalAgent} achieves an 84\% win-tie rate over baselines and 65.0\% Eval@1---the rate at which generated evaluations execute successfully and produce substantive results on the first attempt (\S\ref{sec:main_results}).

Through design space ablations (\S\ref{sec:findings}) and qualitative analysis (\S\ref{sec:qualitative}), we observed:
\begin{enumerate}[leftmargin=*,itemsep=2pt,topsep=3pt]
    \item \textbf{Evaluation skills} stabilize executability as metric count grows (65\% Eval@1 at 5 metrics \vs 30--40\% for baselines), reliably steer generation away from shallow keyword matching toward meaningful semantic assessment, and keep generated code 2.4--6.6$\times$ more concise than baselines.
    \item \textbf{Planning without evaluation domain knowledge hurts more than it helps.} Directly generating evaluation code outperforms planning-then-implementing (74\% win-tie). Unstructured planning expands scope rather than constraining it, producing 4.4$\times$ longer plans, 6.6$\times$ more code, and large amounts of unused files. Planning becomes effective when paired with evaluation skills.
    \item \textbf{\textsc{EvalAgent} is particularly effective under underspecified requirements.} When users provide only generic instructions like ``evaluate this agent'', \textsc{EvalAgent} achieves 95\% win-tie rates over baselines by inferring meaningful metrics from observed behavior, compared to 80\% when requirements are well-specified and simpler methods can partially compensate.
\end{enumerate}

\section{\textsc{EvalAgent}}
\label{sec:method}

\begin{figure*}[t]
\centering
\includegraphics[width=\textwidth]{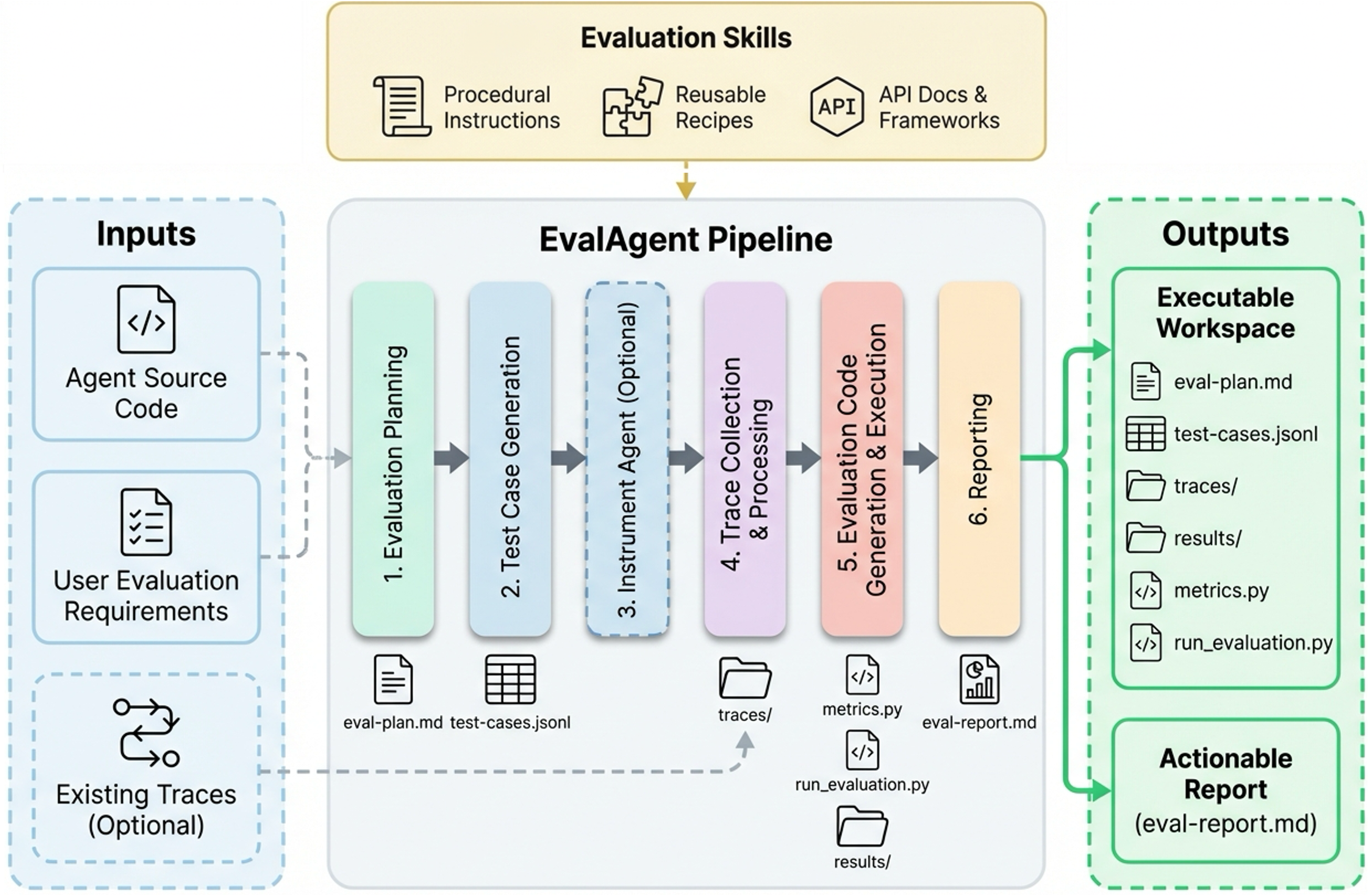}
\caption{\textbf{\textsc{EvalAgent} pipeline.}
\textsc{EvalAgent} maps agent code and user requirements to an executable evaluation workspace and actionable report through a six-stage process: (1)~evaluation planning, (2)~test case generation, (3)~agent instrumentation, (4)~trace collection and processing, (5)~evaluation code generation and execution, and (6)~reporting with actionable recommendations. Evaluation skills guide each stage.
}
\label{fig:system_overview}
\end{figure*}


\textsc{EvalAgent} is an agentic system that transforms agent source code and user requirements into executable evaluation code and actionable reports through a six-stage pipeline (Figure~\ref{fig:system_overview}).

General-purpose coding agents already possess the capabilities needed for evaluation---code understanding, test design, implementation---but lack evaluation-specific knowledge about \emph{what} to measure and \emph{how}. \textsc{EvalAgent} encodes this missing knowledge as \emph{evaluation skills}~\citep{zhang2025agentskills}: self-contained packages of instructions, executable code, and reference materials that the agent loads on demand at each pipeline stage. Skills use progressive disclosure, so only the relevant content is loaded at each stage, and skills are reused across stages. They fall into three categories (Appendix~\ref{app:guidance_examples} provides excerpts):

\textbf{Procedural instructions} encode evaluation methodology as stage-specific workflows. Each skill defines a step-by-step workflow, and constraints that scope the agent's behavior. For instance, the planning skill instructs the agent to first analyze the agent's architecture and trace behavior and identify metrics that each capture a distinct behavioral aspect. The code generation skill prescribes a workflow of implementing metrics, building a main entry point, performing self-review, and updating dependencies---with explicit principles such as creating a minimal working version first, validating library APIs before use, and following the evaluation plan exactly rather than expanding scope.

\textbf{Reusable code and templates} provide implementation scaffolding shared across stages: a plan template with sections for agent specification, metrics, and test scenarios; a report template with executive summary and failure analysis structure; and executable code patterns for OTEL trace parsing and DeepEval~\citep{deepeval2024} metric integration. These reduce variance across runs and prevent the scope expansion observed in unconstrained approaches.

\textbf{Dynamic resources} provide up-to-date external knowledge via Context7~\citep{context7} for real-time API documentation retrieval. Library APIs evolve faster than model training cycles; Context7 fetches current signatures and usage patterns, preventing the deprecated-API errors that account for most executability failures in unconstrained systems (Section~\ref{sec:ablation_context7}).

These skills compose into a six-stage pipeline, where each stage draws on a specific combination of skills (Appendix~\ref{app:pipeline_details} provides full details):

\textbf{(1) Evaluation Planning.} \textsc{EvalAgent} analyzes the agent's source code and execution traces to produce a structured evaluation plan: agent specification, focused metrics with scoring rubrics, and test scenarios.
The planning prompt and plan template jointly guide the scope, preventing metric proliferation.

\textbf{(2) Test Case Generation.} Test cases are generated in JSONL format, each specifying an input query, scenario category, and expected behavior. This phase is substitutable: any generator producing the expected schema can be used.

\textbf{(3) Agent Instrumentation.} Lightweight OpenTelemetry~\citep{opentelemetry_collector} instrumentation is added to the target agent via libraries such as Traceloop~\citep{traceloop}, requiring only a few lines of code.

\textbf{(4) Trace Collection.} The instrumented agent executes test cases, producing OTEL-compatible traces. A trace processor filters to agent-relevant spans and extracts a minimal field set: operation names, inputs/outputs, tool metadata, and timing. The trace extraction code patterns are reused in Phase~5.

\textbf{(5) Code Generation.} \textsc{EvalAgent} translates the evaluation plan into executable Python code, generating metric implementations (both deterministic checks and LLM-based assessments), an evaluation orchestrator, and a dependency manifest. This stage draws on all three skill categories: the code generation prompt enforces simplicity, code patterns provide trace parsing and metric scaffolding, and Context7 retrieves current API signatures.

\textbf{(6) Reporting.} Results are synthesized into a structured report following the report template: executive summary, per-metric analysis, failure root causes, and prioritized recommendations.

\section{Meta-Evaluation and AgentEvalBench}
\label{sec:metaeval}

\begin{figure*}[t]
\centering
\includegraphics[width=\textwidth]{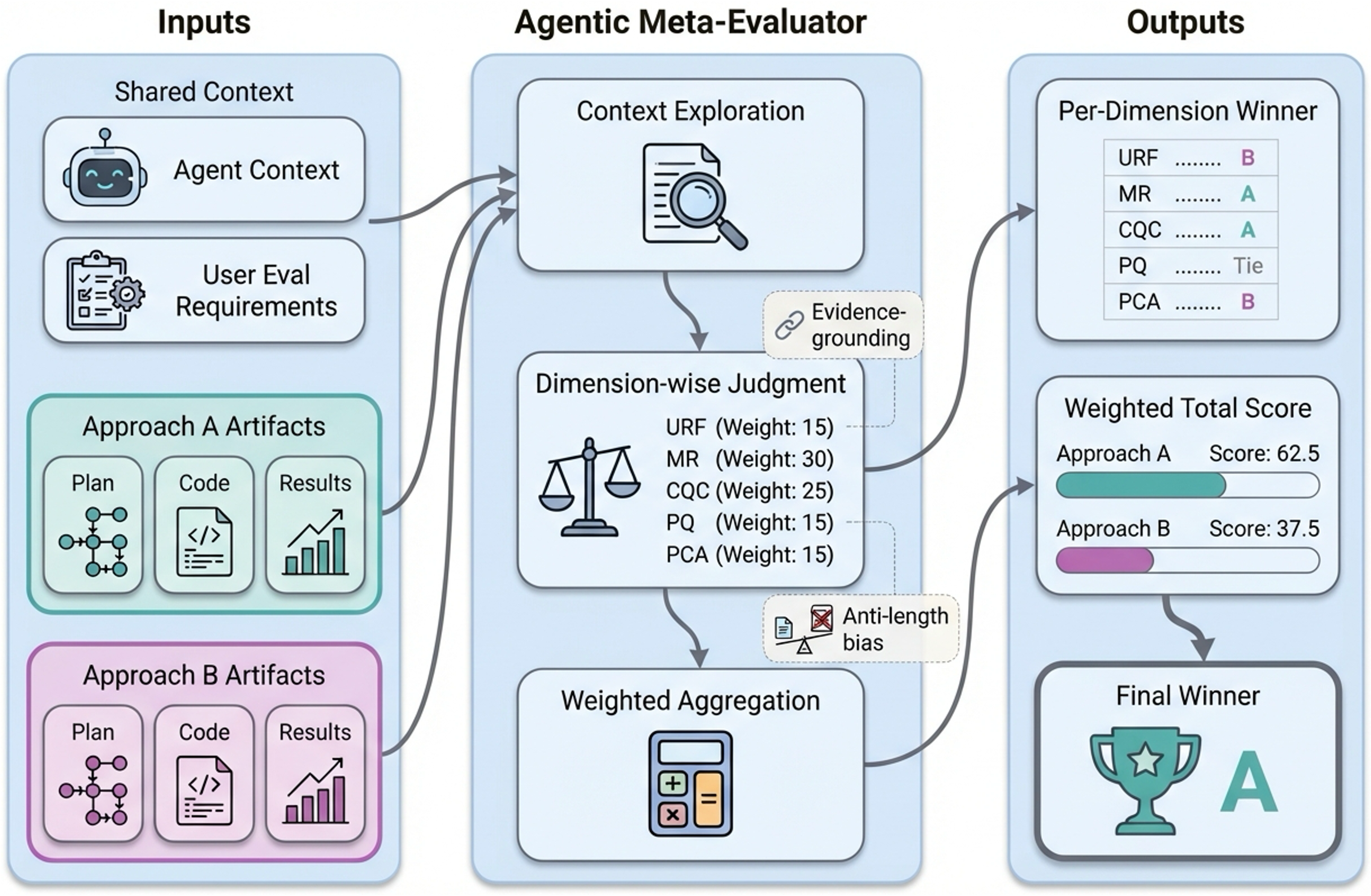}
\caption{\textbf{Meta-evaluation framework.} The agentic meta-evaluator performs pairwise comparisons between two approaches. \textit{(Left)} Shared context and approach-specific artifacts. \textit{(Center)} Evidence-grounded judgment across five weighted criteria with anti-length bias. \textit{(Right)} Weighted aggregation yielding the final winner.}
\label{fig:metaeval_overview}
\end{figure*}

How should we evaluate an evaluator when no ground truth exists? We adopt \emph{comparative validity}: determining which of two evaluation approaches is more useful under identical conditions, following construct validity principles from measurement theory~\citep{cronbach1955construct,messick1995validity}. We further introduce AgentEvalBench, a benchmark for reproducible evaluation of agent evaluation systems.

\subsection{Meta-Evaluation Framework}
\label{subsec:metaeval_framework}
Evaluating an evaluation system is inherently complex: it requires reading multi-file codebases, cross-referencing plans against implementations, and reasoning about whether metrics capture meaningful agent behavior. A standard LLM-as-a-judge approach, which produces verdicts from a single prompt, lacks the capacity for such multi-step inspection. We therefore employ an \emph{agentic} meta-evaluator that judges evaluation artifacts through tool-augmented, multi-step review---navigating directories, reading plans, inspecting code, and examining traces. We adopt pairwise comparison following Thurstone's law of comparative judgment~\citep{thurstone1927law}, since relative judgments are more reliable and require less calibration than absolute scales~\citep{zheng2023judging,wang2024large}. Figure~\ref{fig:metaeval_overview} provides an overview.

\paragraph{Dimensions.}
The meta-evaluator assesses quality across five dimensions (Table~\ref{tab:eval_dimensions}). Dimension weights are set heuristically: Metric Relevance (MR) is highest (30\%) because irrelevant metrics yield no actionable insight regardless of implementation quality; Code Quality \& Complexity (CQC) follows (25\%), as concise yet correct is necessary for practical adoption. The remaining weight is split equally among User Requirement Fulfillment (URF), Plan Quality (PQ), and Plan-Code Alignment (PCA) (15\% each), which capture alignment with user intent and internal consistency. We explicitly enforce an \emph{anti-length bias}: verbosity is not a proxy for quality. When plans are unavailable (\ie, for methods without a planning stage), weights renormalize to URF (25\%), MR (40\%), CQC (35\%).


\begin{table}[htbp]
\centering
\footnotesize
\setlength{\tabcolsep}{4pt}
\renewcommand{\arraystretch}{1.15}
\begin{tabular}{l c p{10cm}}
\toprule
Dimension & Weight & Description \\
\graycmidrule{1-3}
User Requirement Fulfillment (URF) & 15\% & Which approach better addresses user requirements? \\
Metric Relevance (MR) & 30\% & Which has a higher signal-to-noise ratio in its metrics? \\
Code Quality \& Complexity (CQC) & 25\% & Which has more correct, organized, and maintainable code? \\
Plan Quality (PQ) & 15\% & Which plan is more coherent, complete, and actionable? \\
Plan-Code Alignment (PCA) & 15\% & Which implementation more faithfully follows its plan? \\
\bottomrule
\end{tabular}
\caption{\textbf{Meta-evaluation dimensions.} Each dimension yields A wins, B wins, or Tie. When plans are unavailable, weights renormalize to URF (25\%), MR (40\%), CQC (35\%). See Appendix~\ref{app:rubrics} for detailed rubrics.}
\label{tab:eval_dimensions}
\end{table}

\paragraph{Scoring.}
For each dimension, the meta-evaluator assigns A wins, B wins, or Tie with evidence. Overall scores use weighted aggregation:
\begin{equation}
    \text{Points}_A = \sum_{d \in \mathcal{D}} w_d \cdot \mathbf{1}[\text{A wins } d] + 0.5 \cdot w_d \cdot \mathbf{1}[\text{Tie on } d].
\end{equation}
Our meta-evaluation focuses on plan construction and code synthesis using pre-collected traces to ensure controlled comparison.

\paragraph{Evaluation success rate.}
Beyond qualitative meta-evaluation, we measure \textbf{Eval@1} (evaluation success rate): the fraction of agents for which generated evaluation code both executes without errors \emph{and} produces meaningful, non-vacuous results on the first run. Unlike the standard Pass@1 metric used in code generation~\citep{chen2021evaluating}, which only checks whether code runs, Eval@1 requires that the evaluation actually performs a valid evaluation. For example, we count as failures evaluations that (i)~run but produce uninformative outputs (\eg, constant or all-zero metrics), (ii)~run on synthetic, mocked, or hardcoded data instead of real agent artifacts, or (iii)~treat predicted values as the gold standard, which always gives the same score regardless of the agent’s behavior.

\subsection{AgentEvalBench}
\label{subsec:agentevalbench}
We introduce AgentEvalBench, to our knowledge the first benchmark for automated agent evaluation. It comprises 20 real-world agents, each paired with 5 test scenarios and 2 types of user requirements.

\paragraph{Dataset.}
Agents were selected along three axes: (1)~coverage of core capabilities (QA, RAG, tool use, multi-agent coordination), (2)~representation of major frameworks alongside bespoke implementations, and (3)~balanced complexity distribution across simple, medium, and hard tiers. Table~\ref{tab:agent_inventory} summarizes the full inventory. Agents expose up to 9 tools (mean 3.2), with traces ranging from 34 to 350+ events.

\paragraph{Test scenarios.}
Each agent is paired with 5 test scenarios. For instance, \texttt{career\_assist} scenarios cover job search, resume optimization, interview preparation, learning paths, and career transitions, while \texttt{network\_switch\_operator\_agent} scenarios include device statistics queries and configuration tasks (see more examples in Appendix~\ref{sec:example-scenarios}).

\paragraph{User requirements.}
Each agent has two types of evaluation requirements (see Appendix~\ref{sec:eval-req-examples} for examples). (i) Generic: ``Given the agent codebase and execution traces, create an evaluation code.'' This tests whether evaluation methods can infer appropriate metrics without domain guidance. (ii) Specific: \eg, ``evaluate correctness of extracted medical entities'' for \texttt{medical\_document\_processor} or ``precision and recall of recommendations'' for \texttt{agent4rec\_agent}.

\begin{table}[t]
\centering
\footnotesize
\setlength{\tabcolsep}{3pt}
\renewcommand{\arraystretch}{1.15}
\begin{adjustbox}{max width=\linewidth}
\begin{tabular}{l l c r r r c c l}
\toprule
Agent Name & Framework & Complexity & LOC & \#Tools & Trace Len. & Memory & Multi-Agent & Domain \\
\graycmidrule{1-9}
adala\_agent & Custom-Adala & Hard & 3,900 & 9 & 156 & \cmark & \xmark & Data Labeling \\
agent4rec\_agent & LangChain & Hard & 5,825 & 3 & 89 & \cmark & \xmark & Recommendation \\
ai\_tic\_tac\_toe\_agent & Agno & Simple & 369 & 0 & 45 & \xmark & \cmark & Game Playing \\
airbnb\_mcp\_agent & MCP/Bedrock & Simple & 257 & 1 & 112 & \xmark & \xmark & Travel Assistant \\
browser\_mcp\_agent & MCP/Bedrock & Simple & 330 & 4 & 134 & \xmark & \xmark & Web Browsing \\
career\_assist & LangChain/LangGraph & Simple & 383 & 5 & 287 & \xmark & \xmark & Career Planning \\
chat\_with\_pdf\_agent & Embedchain & Simple & 214 & 2 & 78 & \cmark & \xmark & Document QA \\
chat\_with\_research\_papers & Agno & Simple & 203 & 1 & 95 & \xmark & \xmark & Research Assistant \\
city\_agent & LangGraph & Medium & 589 & 4 & 52 & \xmark & \xmark & Information Retrieval \\
cleverchatty\_agent & MCP & Medium & 721 & 3 & 67 & \cmark & \xmark & Conversational AI \\
code\_assistant & Strands & Medium & 567 & 5 & 198 & \xmark & \xmark & Code Generation \\
data\_warehouse\_optimizer & Strands & Medium & 607 & 4 & 167 & \xmark & \cmark & Database Optimization \\
hilton\_search\_agent & LangChain/LangGraph & Hard & 1,513 & 3 & 245 & \cmark & \xmark & Hotel Search \\
llamaindex\_complex\_agent & LlamaIndex & Medium & 641 & 5 & 312 & \cmark & \xmark & Financial Analysis \\
local\_news\_agent & Custom & Simple & 275 & 1 & 63 & \xmark & \cmark & News Aggregation \\
medical\_document\_processor & Strands & Medium & 768 & 8 & 178 & \xmark & \xmark & Medical NLP \\
minisweagent & Custom & Simple & 244 & 0 & 356 & \xmark & \xmark & Software Engineering \\
network\_switch\_operator\_agent & Custom (Bedrock) & Hard & 1,117 & 4 & 89 & \xmark & \xmark & Network Management \\
simple\_qa\_search & LangGraph & Simple & 317 & 1 & 34 & \xmark & \xmark & Question Answering \\
singleagent\_trip\_crew & CrewAI & Simple & 301 & 2 & 234 & \xmark & \xmark & Trip Planning \\
\graycmidrule{1-9}
\textbf{Mean} & -- & -- & 957 & 3.2 & 145 & -- & -- & -- \\
\bottomrule
\end{tabular}
\end{adjustbox}
\caption{\textbf{AgentEvalBench inventory.} 20 agents, 9 frameworks, 14 domains, 3 complexity tiers. LOC excludes comments and blank lines. \#Tools denotes distinct callable tools/functions. Trace Len. is the average number of events per execution trace.}
\label{tab:agent_inventory}
\end{table}

\subsection{Validating the Meta-Evaluator}
\label{subsec:human_validation}

\begin{wraptable}[21]{r}{0.5\linewidth}
\centering
\small
\begin{tabular}{lc}
\toprule
\textbf{Metric} & \textbf{Value} \\
\graycmidrule{1-2}
\multicolumn{2}{l}{\textit{Human annotators ($n{=}3$, \textsc{EvalAgent} \vs B4)}} \\
Prefer \textsc{EvalAgent} & 79.5\% (ties: 10.5\%) \\
Fleiss' $\kappa$ & 0.923 \\
\graycmidrule{1-2}
\multicolumn{2}{l}{\textit{Human \vs Meta-evaluator match}} \\
Overall winner match & 97.5\% (39/40 cases) \\
Per-judgment match & 75.0\% (150/200) \\
\quad Gwet's AC1 & 0.687 \\
\quad Spearman $\rho$ & 0.617 \\
\graycmidrule{1-2}
\multicolumn{2}{l}{\textit{Per-dimension match ($n{=}40$ cases)}} \\
\quad User Requirement Fulfillment & 95.0\% \\
\quad Metric Relevance & 97.5\% \\
\quad Code Quality \& Complexity & 72.5\% \\
\quad Plan Quality & 50.0\% \\
\quad Plan-Code Alignment & 60.0\% \\
\bottomrule
\end{tabular}
\caption{\textbf{Human expert \vs meta-evaluator agreement} on \textsc{EvalAgent} \vs B4. (Top) outcomes from three human annotators. (Middle) meta-evaluator agreement with the human-majority decision for the overall winner and individual dimension judgments (200 total). (Bottom) per-dimension agreement across 40 cases.}
\label{tab:llmj_alignment}
\end{wraptable}

\paragraph{Expert annotator \vs meta-evaluator alignment.}
Table~\ref{tab:llmj_alignment} reports agreement between the meta-evaluator (Claude Opus~4.5) and human experts. Three expert annotators independently performed blinded pairwise comparisons on 40 agent-evaluation pairs (20 agents $\times$ 2 requirement types) across 5 dimensions (200 dimension-level judgments), with randomized presentation order. Experts preferred \textsc{EvalAgent} in 79.5\% of comparisons (ties: 10.5\%, baseline: 10.0\%), with near perfect inter-annotator reliability (Fleiss' $\kappa = 0.923$).
The meta-evaluator closely tracks human judgments, matching the human-majority winner in 97.5\% of cases (39/40) and achieving 75\% agreement at the individual dimension level (Gwet's AC1 = 0.687). Per-dimension match rates are highest for Metric Relevance (97.5\%) and URF (95.0\%), and lowest for Plan Quality (50.0\%) and Plan-Code Alignment (60.0\%)---the latter two reflect inherent subjectivity. Using Sonnet~4.5 as meta-evaluator yields comparable alignment (90.0\% case-level, Gwet's AC1 = 0.696; see Table~\ref{app:llmj_alignment_detail}).

\paragraph{Consistency analysis.}
Beyond human alignment, we assess meta-evaluator reliability through two complementary tests inspired by generalizability theory~\citep{brennan1992generalizability}; detailed results are in Appendix~\ref{app:consistency}.

(i) \emph{Cross-model consistency} tests whether different LLMs serving as meta-evaluators reach the same conclusions on identical evaluation pairs, ensuring that results are not an artifact of a particular model's biases. Using Claude Opus 4.5 and Sonnet 4.5~\citep{anthropic2025claude45sonnet,anthropic2025claude45opus} as meta-evaluators yields 86.6\% overall agreement, with both models preserving the same baseline ranking (Appendix~\ref{app:cross_model}).

(ii) \emph{Run-to-run consistency} tests whether the same meta-evaluator produces stable judgments across independent runs, ruling out stochastic artifacts from sampling. Three independent runs of Claude Opus 4.5 produce 76.3\% three-way agreement and 84.2\% pairwise agreement---well above the 33.3\% random baseline (Appendix~\ref{app:run_consistency}).

\section{Experiments}
\label{sec:experiments}

\paragraph{Setup.}
\label{sec:exp_setup}
Our experiments evaluate the core challenges of automated agent evaluation: plan construction (what to measure) and code synthesis (correct implementation). All methods operate on the same set of pre-collected execution traces.
We use a 2\,$\times$\,2 factorial design: two evaluator backbones (Claude~Haiku~4.5, Sonnet~4.5~\citep{anthropic2025claude45sonnet}) and two requirement types (generic, specific). The benchmark comprises 20 agents from AgentEvalBench. All meta-evaluations in the main paper use Claude Opus~4.5~\citep{anthropic2025claude45opus} as the meta-evaluator backbone; results with Sonnet~4.5 as meta-evaluator are provided in Appendix~\ref{app:sonnet_metaeval} for robustness. \textsc{EvalAgent} is built on Claude Code v2.1.92, with Claude Sonnet 4.5 as its primary backbone, generates evaluation code integrating DeepEval~\citep{deepeval2024}, and queries Context7~\citep{context7} for up-to-date API documentation during code generation.

\paragraph{Baselines.}
We compare against four baselines (see Table~\ref{tab:baselines}), each isolating specific design choices. \textit{LLM-Singleturn (B1)} generates evaluation code in a single prompt: agent code, traces, and requirements are embedded directly in the prompt, with no tool access or iterative refinement. The remaining baselines (B2, B3, B4) and EvalAgent are all built on top of a coding assistant (Claude Code\footnote{We used the same version (v2.1.92) as \textsc{EvalAgent}.}) with full tool access---file reading, code editing, shell execution, and web search---and can perform multi-step reasoning over up to 50 tool calls. \textit{Agent-Sourcecode (B2)} operates on source code only, without execution traces. \textit{Agent-Onestage (B3)} adds trace access for direct code generation. \textit{Agent-Twostage (B4)} introduces an explicit planning stage before implementation but lacks evaluation skills. \textsc{EvalAgent} \textit{(Ours)} combines agentic reasoning, traces, planning, and evaluation skills.

\begin{table}[t]
\centering
\setlength{\tabcolsep}{3pt}
\renewcommand{\arraystretch}{1.15}
\begin{adjustbox}{max width=\linewidth}
\begin{tabular}{c l c l c c c l}
\toprule
Baseline & Name & Approach & Input & Traces & Planning & Skills & Key Characteristics \\
\graycmidrule{1-8}
B1 & LLM-Singleturn & LLM & Code + Traces & \cmark & \xmark & \xmark & Single prompt, no iteration \\
B2 & Agent-Sourcecode & Agentic & Code only & \xmark & \xmark & \xmark & Multi-turn, no trace access \\
B3 & Agent-Onestage & Agentic & Code + Traces & \cmark & \xmark & \xmark & Direct code generation \\
B4 & Agent-Twostage & Agentic & Code + Traces & \cmark & \cmark & \xmark & Plan then implement \\
Ours & \textsc{EvalAgent} & Agentic & Code + Traces & \cmark & \cmark & \cmark & Evaluation skills \\
\bottomrule
\end{tabular}
\end{adjustbox}
\captionsetup{justification=centering}
\caption{\textbf{Baselines.} All agentic methods share identical tool access (\eg, file reading, code execution).}
\label{tab:baselines}
\end{table}

\begin{table*}[t]
\centering
\begin{subtable}{\textwidth}
\centering
\begin{tabular}{c r c c c c c c}
\toprule
& & \multicolumn{6}{c}{\textbf{Win-Tie Rate} (\%)} \\
\graycmidrule{3-8}
LLM & Comparison & \makecell{URF} & \makecell{MR} & \makecell{CQC} & \makecell{PQ} & \makecell{PCA} & \makecell{Overall} \\
\graycmidrulethree{1-1}{2-2}{3-8}
\multirow{6}{*}{\rotatebox{90}{Haiku 4.5}} & Ours \vs B1 & \makecell{85.0\\[-2pt]{\tiny(29W/10T/1L)}} & \makecell{96.2\\[-2pt]{\tiny(38W/1T/1L)}} & \makecell{95.0\\[-2pt]{\tiny(36W/4T/0L)}} & - & - & \makecell{97.4\\[-2pt]{\tiny(38W/1T/1L)}} \\
& \quad \vs B2 & \makecell{80.0\\[-2pt]{\tiny(24W/16T/0L)}} & \makecell{98.8\\[-2pt]{\tiny(39W/1T/0L)}} & \makecell{97.5\\[-2pt]{\tiny(38W/2T/0L)}} & - & - & \makecell{100.0\\[-2pt]{\tiny(40W/0T/0L)}} \\
& \quad \vs B3 & \makecell{68.8\\[-2pt]{\tiny(18W/19T/3L)}} & \makecell{92.5\\[-2pt]{\tiny(37W/0T/3L)}} & \makecell{96.2\\[-2pt]{\tiny(38W/1T/1L)}} & - & - & \makecell{92.5\\[-2pt]{\tiny(37W/0T/3L)}} \\
& \quad \vs B4 & \makecell{60.0\\[-2pt]{\tiny(14W/20T/6L)}} & \makecell{83.8\\[-2pt]{\tiny(31W/5T/4L)}} & \makecell{92.5\\[-2pt]{\tiny(36W/2T/2L)}} & \makecell{75.0\\[-2pt]{\tiny(29W/2T/9L)}} & \makecell{68.8\\[-2pt]{\tiny(21W/13T/6L)}} & \makecell{87.5\\[-2pt]{\tiny(35W/0T/5L)}} \\
\graycmidrule{1-8}
\multirow{6}{*}{\rotatebox{90}{Sonnet 4.5}} & Ours \vs B1 & \makecell{85.0\\[-2pt]{\tiny(32W/4T/4L)}} & \makecell{90.0\\[-2pt]{\tiny(36W/0T/4L)}} & \makecell{88.8\\[-2pt]{\tiny(32W/7T/1L)}} & - & - & \makecell{90.0\\[-2pt]{\tiny(36W/0T/4L)}} \\
& \quad \vs B2 & \makecell{83.8\\[-2pt]{\tiny(30W/7T/3L)}} & \makecell{93.8\\[-2pt]{\tiny(37W/1T/2L)}} & \makecell{96.2\\[-2pt]{\tiny(38W/1T/1L)}} & - & - & \makecell{94.9\\[-2pt]{\tiny(37W/1T/2L)}} \\
& \quad \vs B3 & \makecell{71.2\\[-2pt]{\tiny(24W/9T/7L)}} & \makecell{78.8\\[-2pt]{\tiny(29W/5T/6L)}} & \makecell{81.2\\[-2pt]{\tiny(30W/5T/5L)}} & - & - & \makecell{84.2\\[-2pt]{\tiny(32W/1T/7L)}} \\
& \quad \vs B4 & \makecell{63.8\\[-2pt]{\tiny(15W/21T/4L)}} & \makecell{87.5\\[-2pt]{\tiny(34W/2T/4L)}} & \makecell{92.5\\[-2pt]{\tiny(36W/2T/2L)}} & \makecell{83.8\\[-2pt]{\tiny(32W/3T/5L)}} & \makecell{61.3\\[-2pt]{\tiny(17W/15T/8L)}} & \makecell{90.0\\[-2pt]{\tiny(36W/0T/4L)}} \\
\bottomrule
\end{tabular}
\captionsetup{width=\textwidth}
\caption{
\textbf{Meta-evaluation results (Opus 4.5 as Judge).} PQ/PCA not applicable (-) for B1, B2, B3 (no planning). Win-Tie Rate = (wins + 0.5$\times$ties) / total $\times$ 100; higher is better for \textsc{EvalAgent}; W = wins, T = ties, L = losses for \textsc{EvalAgent}.
}
\label{tab:win_rate}
\end{subtable}%
\vspace{1em}
\begin{subtable}[t]{0.3\textwidth}
\centering
\setlength{\tabcolsep}{4pt}
\begin{tabular}{c c c}
\toprule
LLM & Approach & \textbf{Eval@1} (\%)\\
\graycmidrulethree{1-1}{2-2}{3-3}
\multirow{5}{*}{\rotatebox{90}{Haiku 4.5}} & B1 & 15.0 {\scriptsize(6/40)} \\
 & B2 & 45.0 {\scriptsize(18/40)} \\
 & B3 & 17.5 {\scriptsize(7/40)} \\
 & B4 & 32.5 {\scriptsize(13/40)} \\
 & Ours & 62.5 {\scriptsize(25/40)} \\
\graycmidrule{1-3}
\multirow{5}{*}{\rotatebox{90}{Sonnet 4.5}} & B1 & 17.5 {\scriptsize(7/40)} \\
 & B2 & 60.0 {\scriptsize(24/40)} \\
 & B3 & 35.0 {\scriptsize(14/40)} \\
 & B4 & 30.0 {\scriptsize(12/40)} \\
 & Ours & 65.0 {\scriptsize(26/40)} \\
\bottomrule
\end{tabular}
\captionsetup{justification=centering}
\caption{\textbf{Eval success rate.}}
\label{tab:code_exec}
\end{subtable}\hfill
\begin{subtable}[t]{0.4\textwidth}
\centering
\setlength{\tabcolsep}{4pt}
\begin{tabular}{c c c c}
\toprule
LLM & Approach & \makecell{\textbf{Time} (min)} & \makecell{\textbf{Tokens} (K)} \\
\graycmidrulethree{1-1}{2-2}{3-4}
\multirow{5}{*}{\rotatebox{90}{Haiku 4.5}} & B1 & 0.68 & 50.26 \\
& B2 & 4.54 & 2196.16 \\
& B3 & 5.30 & 3579.46 \\
& B4 & 6.72 & 4049.63 \\
& Ours & 3.98 & 2872.07 \\
\graycmidrule{1-4}
\multirow{5}{*}{\rotatebox{90}{Sonnet 4.5}} & B1 & 1.36 & 49.99 \\
& B2 & 4.03 & 869.36 \\
& B3 & 4.39 & 1698.44 \\
& B4 & 10.00 & 3023.59 \\
& Ours & 4.18 & 2094.98 \\
\bottomrule
\end{tabular}
\captionsetup{justification=centering}
\caption{\textbf{Efficiency.}}
\label{tab:cost_analysis}
\end{subtable}\hfill
\begin{minipage}{0.3\textwidth}
\vspace{0pt}
\centering
\resizebox{0.95\linewidth}{!}{%
\begin{tabular}{rl}
\toprule
\multicolumn{2}{c}{\textbf{Legend}} \\
\midrule
B1 & LLM-Singleturn \\
B2 & Agent-Sourcecode \\
B3 & Agent-Onestage \\
B4 & Agent-Twostage \\
\graycmidrule{1-2}
URF & User Requirement Fulfillment \\
MR & Metric Relevance \\
CQC & Code Quality \& Complexity \\
PQ & Plan Quality \\
PCA & Plan-Code Alignment \\
\bottomrule
\end{tabular}%
}
\end{minipage}
\caption{\textbf{Main results.}
\textbf{(a)} Win-tie rates across 5 meta-evaluation dimensions.
\textbf{(b)} Eval@1 (evaluation success rate) across 20 agents $\times$ 2 requirement types.
\textbf{(c)} Efficiency. Time = pure execution time to generate evaluations excluding network latency or API delays; Tokens = total tokens consumed.
}
\label{tab:main_comparison}
\end{table*}

\subsection{Main Results}
\label{sec:main_results}

Table~\ref{tab:main_comparison} presents the complete comparison across meta-evaluation quality, evaluation success rate, and efficiency.

\paragraph{Meta-evaluation results.}
\label{sec:meta_eval_results}
\textsc{EvalAgent} achieves 84--100\% overall win-tie rates across all baselines and both evaluator LLMs (Table~\ref{tab:win_rate}). The most informative comparison is against B4, which shares the same two-stage architecture but lacks evaluation skills: \textsc{EvalAgent} achieves 87.5--90.0\% overall win-tie, with particularly strong advantages in Code Quality \& Complexity (92.5\%) and Metric Relevance (83.8--87.5\%). This demonstrates that evaluation skills, not just planning, drive quality gains. Against weaker baselines, advantages are even larger: B1 and B2 lack either iterative reasoning or trace access, yielding 90--100\% win-tie rates for \textsc{EvalAgent}.

\paragraph{Evaluation success rate.}
Table~\ref{tab:code_exec} reports Eval@1. B1 achieves only 15.0--17.5\%, confirming that single-pass generation is insufficient. B2 achieves the highest Eval@1 (45.0--60.0\%) among baselines because source-code-only evaluation avoids trace parsing complexity. \textsc{EvalAgent} achieves 62.5--65.0\%, substantially outperforming all baselines, including B4 (30.0--32.5\%) which shares the same two-stage architecture but lacks evaluation skills. The stronger Sonnet backbone generally improves Eval@1 across methods.

\paragraph{Efficiency.}
\label{sec:cost_efficiency}
\textsc{EvalAgent} achieves a favorable quality--cost trade-off (Table~\ref{tab:cost_analysis}). Compared to B4, which shares the same two-stage architecture, it uses 31\% fewer tokens (2,095K \vs 3,024K for Sonnet) and 58\% less time (4.18 \vs 10.00\,min) while achieving higher quality. This demonstrates that evaluation skills reduce unnecessary exploration overhead. B1 is the most token-efficient but produces the lowest quality.

\subsection{Findings}
\label{sec:findings}

\paragraph{Finding 1: \textsc{EvalAgent} is robust under both generic and specific requirements.}
\label{sec:performance_specificity}
\begin{wraptable}[12]{r}{0.6\linewidth}
\centering
\setlength{\tabcolsep}{4pt}
\renewcommand{\arraystretch}{1.15}
\vspace{-3mm}
\begin{subtable}{0.49\linewidth}
\centering
\begin{tabular}{r c c}
\toprule
\multicolumn{3}{c}{\textbf{Win-Tie Rate by Requirement}} \\
Comparison & Generic & Specific  \\
\graycmidruletwo{1-1}{2-3}

Ours \vs B1 & 90.0\% & 97.4\% \\
\quad \vs B2 & 100.0\% & 94.9\% \\
\quad \vs B3 & 95.0\% & 81.6\% \\
\quad \vs B4 & 97.4\% & 79.5\% \\
\bottomrule
\end{tabular}
\captionsetup{justification=centering}
\caption{\textbf{Meta-evaluation.}}
\label{tab:specificity_meta}
\end{subtable}%
\hfill
\begin{subtable}{0.49\linewidth}
\centering
\setlength{\tabcolsep}{4pt}
\begin{tabular}{c c c}
\toprule
\multicolumn{3}{c}{\textbf{Eval@1 by Requirement}} \\
Approach & Generic & Specific \\
\graycmidruletwo{1-1}{2-3}

B1 & 7.5\% & 25.0\% \\
B2 & 50.0\% & 55.0\% \\
B3 & 20.0\% & 32.5\% \\
B4 & 30.0\% & 32.5\% \\
Ours & 65.0\% & 62.5\% \\
\bottomrule
\end{tabular}
\captionsetup{justification=centering}
\caption{\textbf{Eval success rate.}}
\label{tab:specificity_exec}
\end{subtable}
\caption{
\textbf{Performance by requirement type.} Results aggregated across Haiku 4.5 and Sonnet 4.5.
}
\label{tab:specificity_analysis}
\end{wraptable}
As Table~\ref{tab:specificity_analysis} shows, \textsc{EvalAgent} maintains strong performance across both requirement types, achieving 62.5--65.0\% Eval@1 regardless of specificity. Its advantage is especially pronounced under generic requirements (``evaluate this agent'' without detailed specifications), where the system must infer appropriate metrics from agent behavior alone: 95.0\% win-tie against B3 and 97.4\% against B4. When requirements are well-specified, baselines partially close the gap (B3 rises from 20.0\% to 32.5\% Eval@1), but \textsc{EvalAgent} retains a substantial lead (62.5\% \vs 32.5\%). This robustness across both conditions reflects the benefit of evaluation skills, which provide principled defaults when user guidance is minimal and complement domain-specific instructions when available.

\paragraph{Finding 2: Traces improve quality but reduce executability.}
\begin{wraptable}[9]{r}{0.6\linewidth}
\centering
\setlength{\tabcolsep}{4pt}
\renewcommand{\arraystretch}{1.15}
\vspace{-3mm}
\begin{subtable}{0.6\linewidth}
\centering
\begin{tabular}{c c c c c}
\toprule
\multicolumn{5}{c}{\textbf{Win-Tie Rate (B3 \vs B2)} (\%)} \\
LLM & \makecell{URF} & \makecell{MR} & \makecell{CQC} & \makecell{Overall} \\
\graycmidruletwo{1-1}{2-5}

Haiku 4.5 & 67.1 & 80.3 & 72.4 & 73.2 \\
Sonnet 4.5 & 70.0 & 80.0 & 80.0 & 76.7 \\
\bottomrule
\end{tabular}
\captionsetup{justification=centering}
\caption{\textbf{Meta-evaluation.}}
\label{tab:win_rate_3vs2}
\end{subtable}%
\begin{subtable}{0.39\linewidth}
\centering
\begin{tabular}{c c c}
\toprule
& \multicolumn{2}{c}{\textbf{Eval@1} (\%)} \\
LLM & B2 & B3 \\
\graycmidruletwo{1-1}{2-3}

Haiku 4.5 & 45.0 & 17.5 \\
Sonnet 4.5 & 60.0 & 35.0 \\
\bottomrule
\end{tabular}
\captionsetup{justification=centering}
\caption{\textbf{Eval success rate.}}
\label{tab:code_exec_3vs2}
\end{subtable}
\caption{
\textbf{B3 \vs B2: trace-based \vs source-code-only.} Win-tie rates are reported for B3 over B2.
}
\label{tab:3vs2}
\end{wraptable}
Trace-based evaluation (B3) achieves 73--77\% win-tie rates over source-code-only (B2), with the strongest advantage in Metric Relevance (80\%). Traces enable metrics that assess \emph{what the agent actually did}---runtime tool calls, error recovery, and decision sequences invisible to static analysis. However, trace parsing introduces complexity that reduces Eval@1: B2 achieves 60.0\% \vs B3's 35.0\% with Sonnet. Part of this gap stems from the inherent difficulty of parsing traces, but B2's advantage is also inflated by evaluations that run on unit test stubs rather than real agent behavior. While trace-based evaluation imposes higher implementation overhead, it remains essential for capturing behavioral failures and decision-level errors that static code analysis cannot detect.

\paragraph{Finding 3: Planning without structure hurts quality.}
\begin{wraptable}[8]{r}{0.6\linewidth}
\centering
\setlength{\tabcolsep}{4pt}
\renewcommand{\arraystretch}{1.15}
\vspace{-3mm}
\begin{subtable}{0.6\linewidth}
\centering
\begin{tabular}{c c c c c}
\toprule
\multicolumn{5}{c}{\textbf{Win-Tie Rate (B3 \vs B4)} (\%)} \\
LLM & \makecell{URF} & \makecell{MR} & \makecell{CQC} & \makecell{Overall} \\
\graycmidruletwo{1-1}{2-5}

Haiku 4.5 & 52.6 & 56.4 & 60.3 & 56.4 \\
Sonnet 4.5 & 57.5 & 78.8 & 85.0 & 73.8 \\
\bottomrule
\end{tabular}
\captionsetup{justification=centering}
\caption{\textbf{Meta-Evaluation.}}
\label{tab:win_rate_3vs4}
\end{subtable}%
\begin{subtable}{0.39\linewidth}
\centering
\begin{tabular}{c c c}
\toprule
& \multicolumn{2}{c}{\textbf{Eval@1} (\%)} \\
LLM & B3 & B4 \\
\graycmidruletwo{1-1}{2-3}

Haiku 4.5 & 17.5 & 32.5 \\
Sonnet 4.5 & 35.0 & 30.0 \\
\bottomrule
\end{tabular}
\captionsetup{justification=centering}
\caption{\textbf{Eval success rate.}}
\label{tab:code_exec_3vs4}
\end{subtable}
\caption{
\textbf{B3 (direct code implementation) \vs B4 (plan-then-implement).} Win-tie rates are reported for B3 over B4.
}
\label{tab:3vs4}
\end{wraptable}
Planning does not automatically improve evaluation quality. With Sonnet, the simpler one-stage approach (B3) achieves a 73.8\% win-tie rate against two-stage (B4), with strong advantages in Code Quality \& Complexity (85.0\%) and Metric Relevance (78.8\%). B4 achieves 30.0--32.5\% Eval@1---comparable to B3's 17.5--35.0\%---because B4's planning stage produces code with broader exception handling that avoids crashes but frequently yields all-zero metrics due to trace format mismatches. The takeaway is that planning requires structure to be effective. Without it, the planning stage becomes an opportunity for scope creep rather than focused design.

\paragraph{Finding 4: Evaluation skills keep executability stable as complexity grows.}

\begin{wraptable}[8]{r}{0.5\linewidth}
\centering
\setlength{\tabcolsep}{5pt}
\renewcommand{\arraystretch}{1.15}
\vspace{-3mm}
\begin{tabular}{c c ccc}
\toprule
& \multicolumn{4}{c}{\textbf{Eval@1 by Number of Metrics}} \\
LLM & Approach & 1 metric & 3 metrics & 5 metrics \\
\graycmidrulethree{1-1}{2-2}{3-5}

\multirow{3}{*}{Sonnet 4.5} & B3 & 55.0\% & 50.0\% & 30.0\% \\
& B4 & 50.0\% & 55.0\% & 40.0\% \\
& Ours & 60.0\% & 55.0\% & 65.0\% \\
\bottomrule
\end{tabular}
\captionsetup{justification=centering}
\caption{\textbf{Eval@1 vs.\ metric count.}}
\label{tab:ablation_num_metrics_exec}
\end{wraptable}
To isolate the effect of evaluation complexity, we fix the prompt to ``implement evaluation code with $k$ metrics'' for $k \in \{1, 3, 5\}$ and measure Eval@1. At 1~metric, all methods perform similarly (55--60\%). As $k$ increases, baseline performance degrades: B3 drops to 30\% at 5~metrics, B4 to 40\%, while \textsc{EvalAgent} improves to 65\%. Each additional metric introduces new trace-parsing logic and cross-metric dependencies. Without skills to provide reusable extraction patterns, failure probability compounds multiplicatively. Evaluation skills factor out shared trace parsing. Since real-world evaluations rarely involve a single metric, this scaling property is practically important.

\paragraph{Finding 5: Access to up-to-date API docs is critical for executability.}
\label{sec:ablation_context7}
\begin{wraptable}[8]{r}{0.3\linewidth}
\centering
\setlength{\tabcolsep}{5pt}
\renewcommand{\arraystretch}{1.15}
\vspace{-2mm}
\begin{tabular}{c cc}
\toprule
& \multicolumn{2}{c}{\textbf{Eval@1}} \\
LLM & w/ C7 & w/o C7 \\
\graycmidruletwo{1-1}{2-3}

Haiku 4.5 & 62.5\% & 47.5\% \\
Sonnet 4.5 & 65.0\% & 20.0\% \\
\bottomrule
\end{tabular}
\captionsetup{justification=centering}
\caption{\textbf{Context7 ablation.}}
\label{tab:ablation_context7_exec}
\end{wraptable}
Context7 improves Sonnet's Eval@1 from 20.0\% to 65.0\%---a 45pp gain. Without it, generated code uses incorrect model identifiers that fail with DeepEval/LiteLLM/Bedrock (see Codes~\ref{alg:context7_without_career}--\ref{alg:context7_with_career}). Context7 provides current documentation showing the correct format and proper class inheritance patterns. Library APIs evolve faster than model training cycles, making dynamic documentation retrieval essential for any code generation system targeting evaluation frameworks.

\subsection{Qualitative Analysis}
\label{sec:qualitative}
We compare \textsc{EvalAgent} against all four baselines on all 20 agents (generic requirements, Sonnet~4.5). Table~\ref{tab:comparison_summary} reports aggregate statistics; Appendix~\ref{app:code_examples} provides concrete examples including directory structures (Figure~\ref{fig:dir_comparison}) and code comparisons (Figure~\ref{fig:code_comparison}).

\begin{table*}[t]
\centering
\setlength{\tabcolsep}{4pt}
\renewcommand{\arraystretch}{1.15}
\small
\begin{tabular}{l c c c c c}
\toprule
& \textbf{\textsc{EvalAgent}} & \textbf{B1} & \textbf{B2} & \textbf{B3} & \textbf{B4} \\
& {\scriptsize Ours} & {\scriptsize LLM-Singleturn} & {\scriptsize Agent-Sourcecode} & {\scriptsize Agent-Onestage} & {\scriptsize Agent-Twostage} \\
\graycmidrule{1-6}
LOC (mean $\pm$ std) & 289 $\pm$ 67 & 694 $\pm$ 85 & 1{,}068 $\pm$ 454 & 844 $\pm$ 271 & 1{,}902 $\pm$ 622 \\
\quad \textit{ratio \vs} \textsc{EvalAgent} & -- & 2.4$\times$ & 3.7$\times$ & 2.9$\times$ & 6.6$\times$ \\
Files (mean $\pm$ std) & 2.0 $\pm$ 0.2 & 3.4 $\pm$ 1.4 & 5.1 $\pm$ 1.2 & 5.4 $\pm$ 2.1 & 8.6 $\pm$ 5.4 \\
Metric classes (mean $\pm$ std) & 1.4 $\pm$ 1.2 & 4.9 $\pm$ 3.6 & 7.1 $\pm$ 4.6 & 7.2 $\pm$ 3.3 & 12.0 $\pm$ 5.6 \\
Plan lines (mean $\pm$ std) & 134 $\pm$ 9 & -- & -- & -- & 584 $\pm$ 264 \\
\graycmidrule{1-6}
\multicolumn{6}{l}{\textit{Evaluation strategy (agents out of 20):}} \\
\quad Keyword heuristics only & 0 & 2 & 3 & 5 & 5 \\
\quad LLM-as-Judge only & 13 & 4 & 5 & 7 & 4 \\
\quad Both & 4 & 10 & 9 & 2 & 10 \\
\quad Neither & 3 & 4 & 3 & 6 & 1 \\
\graycmidrule{1-6}
\multicolumn{6}{l}{\textit{Dead-code artifacts (out of 20 agents):}} \\
\quad \path{report_generator.py} & 0 & 0 & 1 & 3 & 11 \\
\quad \path{test_*.py} (never executed) & 0 & 3 & 19 & 12 & 15 \\
\quad \path{ground_truth_generator.py} & 0 & 0 & 0 & 0 & 3 \\
\bottomrule
\end{tabular}
\caption{\textbf{Artifact comparison across all 20 agents} (generic requirements, Sonnet~4.5). All baselines produce 2.4--6.6$\times$ more code than \textsc{EvalAgent}. Dead-code artifacts accumulate in agentic baselines (B2--B4), with B2 generating unexecuted test files in 19/20 agents. B4's plans are 4.4$\times$ longer than \textsc{EvalAgent}'s yet less actionable (Insight~2).}
\label{tab:comparison_summary}
\end{table*}

\paragraph{Insight 1: Without evaluation skills, baselines default to keyword matching instead of semantic evaluation.}
Keyword heuristics---where the generated evaluation code scores agent outputs by counting regex or substring matches against hardcoded word lists---appear in 35--75\% of baseline agents (Table~\ref{tab:comparison_summary}); \textsc{EvalAgent} reduces this to 20\%. For example, B4's \texttt{singleagent\_trip\_crew} scores itineraries via regex against 30 hardcoded action verbs, so ``the museum was closed'' scores identically to a detailed visit plan (Figure~\ref{fig:code_comparison}). \textsc{EvalAgent} instead uses LLM-as-Judge (GEval) with structured criteria. Neither agentic reasoning, trace access, nor planning alone reliably shifts baselines from keyword matching to semantic evaluation; only evaluation skills do.

\paragraph{Insight 2: Code complexity scales with autonomy level, but evaluation quality does not.}
\textsc{EvalAgent} produces the least code across all 20 agents (289~LOC), while baselines generate 2.4--6.6$\times$ more (Table~\ref{tab:comparison_summary}). Crucially, this extra code does not translate to better evaluation strategy. Adding traces to the agentic baseline (B2$\to$B3) actually \emph{worsens} evaluation quality, as the agent shifts toward parsing durations and call counts instead of semantic assessment. Planning (B3$\to$B4) partially recovers strategy but at 2.3$\times$ the code cost. Only \textsc{EvalAgent} achieves both the highest LLM-as-Judge adoption rate (85\%) and the lowest complexity.

\paragraph{Insight 3: Unstructured planning produces unused files instead of constraining scope.}
Dead code---generated files that are never executed or referenced by the evaluation entry point---scales with autonomy (Table~\ref{tab:comparison_summary}). B4 produces the most: 29 dead-code instances across 20 agents, including 11 unused report generators. Planning does not constrain scope---it \emph{licenses} it. For example, B4's \texttt{career\_assist} plan (1{,}031 lines) specifies ``SME review on 1--5 scale''; the code substitutes \texttt{check\_keyword\_presence()}. \textsc{EvalAgent}'s plan (134 lines) specifies 3 metrics with explicit methods, and the code implements exactly those three---yielding a consistent 2-file pattern across 19/20 agents with zero dead code.

\subsection{Failure Analysis}
\label{sec:failure_analysis}

We categorize all execution failures from \textsc{EvalAgent} across both backbones: Haiku~4.5 (15/40 failures, 62.5\% Eval@1) and Sonnet~4.5 (14/40 failures, 65.0\% Eval@1). Table~\ref{tab:failure_analysis} summarizes error categories.

\begin{table}[h]
\centering
\footnotesize
\setlength{\tabcolsep}{4pt}
\renewcommand{\arraystretch}{1.15}
\begin{tabular}{l c c p{5.5cm}}
\toprule
Error Category & Haiku & Sonnet & Description \\
\graycmidrulefour{1-1}{2-2}{3-3}{4-4}
DeepEval async mode & 0 & 9 & Missing \texttt{a\_measure()} for async execution \\
Type handling (timestamps) & 5 & 0 & String timestamps not converted to int \\
Python API misuse & 6 & 0 & Invalid \texttt{dict(os.stat())}, \texttt{Path.ctime()}, etc.\ \\
DeepEval API misuse & 2 & 2 & Wrong constructor / attribute access \\
DeepEval type mismatch & 0 & 2 & Dicts passed where strings expected \\
Import / syntax errors & 2 & 0 & Relative vs.\ absolute imports, indentation \\
Trace extraction logic & 0 & 1 & Empty \texttt{actual\_output} from trace parsing \\
\graycmidrule{1-4}
\textbf{Total failures} & \textbf{15} & \textbf{14} & \\
\bottomrule
\end{tabular}
\caption{\textbf{Execution failure categorization.} Haiku and Sonnet achieve similar pass rates but fail for \emph{disjoint reasons}: Sonnet is dominated by DeepEval async mode (64\%); Haiku by Python API misuse (40\%) and timestamp handling (33\%). Only DeepEval API misuse is shared (2 each).}
\label{tab:failure_analysis}
\end{table}

\paragraph{Model capability affects error types, not error rates.}
Sonnet's failures concentrate on a single systematic issue: DeepEval defaults to \texttt{async\_mode=True}, but Sonnet implements only synchronous \texttt{measure()} without the required \texttt{a\_measure()} coroutine. This accounts for 9/14 failures (64\%); the generated code is structurally correct in every case and works when \texttt{async\_mode=False} is set manually. Haiku's failures are heterogeneous: Python API misuse (6 cases), timestamp string arithmetic (5 cases), plus import and syntax errors entirely absent from Sonnet. Haiku avoids the async trap because it generates simpler code that less frequently subclasses \texttt{BaseMetric}.

\paragraph{Failures are condition-dependent, not agent-dependent.}
For both backbones, most failing agents fail in only \emph{one} requirement condition: Sonnet has 3 agents failing in both conditions vs.\ 8 in only one; Haiku shows the same 3-vs-9 split. Moreover, agents that fail in both conditions often fail for \emph{different reasons}---\eg, \texttt{hilton\_search\_agent} fails on type mismatch (generic) but async mode (specific). This suggests retry-based approaches may be effective, as re-generation would likely avoid the specific edge case.

\paragraph{Implications.}
Fixing Sonnet's async mode alone (a single parameter or 3-line wrapper) would raise Eval@1 from 65\% to an estimated 87.5\%---the highest-leverage intervention. Haiku's syntax and API errors (53\% of its failures) indicate a minimum capability threshold for evaluation code generation that smaller models may not meet. Context7 documentation retrieval is necessary but insufficient: framework-related errors account for 11/14 Sonnet and 2/15 Haiku failures despite up-to-date documentation access.
\section{Related Work}
\label{sec:related_work}
\paragraph{LLM judges.}
\label{sec:related:llm-eval}
LLM-as-a-Judge~\citep{fu2024gptscore,zhu2023judgelm,kim2023prometheus} is now a standard alternative to human evaluation. \citet{zheng2023judging} show that LLM judges can achieve over 80\% agreement with humans; G-Eval~\citep{liu2023geval} improves correlation via chain-of-thought prompting; ChatEval~\citep{chan2023chateval} employs multi-LLM debate to mimic human panel discussions. However, most LLM judges focus on single outputs and are therefore ill-suited to evaluate multi-step traces or agent behavior. 

\paragraph{Agentic judges.}
Agent-as-a-Judge~\citep{zhuge2024agent} enables evaluators to execute code and browse documentation, but relies on manual task annotations and produces binary judgments. MAJ-Eval~\citep{chen2025multi} orchestrates multi-persona debates; DeepResearchEval~\citep{wang2026deepresearcheval} generates persona-driven tasks with adaptive fact-checking; CodeVisionary~\citep{wang2025agent} distills multi-dimensional context from code requirements and produces fine-grained scores. These approaches use agents to \emph{directly judge} target agents. In contrast, \textsc{EvalAgent} tackles a different and harder challenge: It dynamically \textit{generates} agent-specific evaluation pipelines---criteria, metrics, and executable code---to assess execution quality.

\paragraph{Agent benchmarks.}
AgentBench~\citep{liu2024agentbench} evaluates agents across environments including OS commands, databases, knowledge graphs, and web browsing. WebArena~\citep{zhou2023webarena}, Mind2Web~\citep{deng2023mind2web}, and VisualWebArena~\citep{koh2024visualwebarena} assess agents in web environments. SWE-Bench~\citep{jimenez2024swebench} focuses on resolving GitHub issues, and ToolLLM~\citep{qin2023toolllm}, ToolEmu~\citep{ruan2023identifying}, and API-Bank~\citep{li2023api} benchmark tool-use capabilities. Other benchmarks target real computer tasks~\citep{xie2024osworld}, web access~\citep{mialon2023gaia}, workplace workflows~\citep{xu2024theagentcompany}, and scientific discovery~\citep{chen2024scienceagentbench}. These outcome-only evaluations overlook agent dynamics. TRACE~\citep{kim2025beyond} evaluates agent trajectories using an evidence bank. TRAJECT-Bench~\citep{he2025traject} assesses fine-grained tool usage. These benchmarks assess agents against predefined objectives or reference trajectories. In contrast, \textsc{EvalAgent} provides trace-aware, reference-free evaluation across diverse agent types and domains.

\paragraph{Agent evaluation frameworks.}
Several frameworks support agent evaluators through predefined, standardized metrics. DeepEval~\citep{deepeval2024} provides modular metrics with custom LLM-as-judge support. RAGAS~\citep{es2024ragas} offers reference-free evaluation for retrieval-augmented generation. Langfuse~\citep{langfuse} models agents as trajectory loops supporting black-box, glass-box, and white-box evaluations. LangSmith~\citep{langsmith} enables evaluation via curated datasets and pluggable evaluators. These frameworks provide useful building blocks but require manual configuration per agent---selecting metrics, writing instrumentation code, and designing evaluation logic. \textsc{EvalAgent} automates this full pipeline end-to-end.

\paragraph{Meta-evaluation of judges.}
\label{sec:related:judge-eval}
As LLM and agent judges become common, understanding their reliability is increasingly important. Prior work~\citep{zheng2023judging} evaluates judges via correlation with human annotations, robustness to perturbations, and calibration studies. \textsc{JudgeBench}~\citep{tan2024judgebench} benchmarks LLM judges on challenging response pairs across knowledge, reasoning, and coding tasks. Our meta-evaluation framework builds on pairwise comparison~\citep{thurstone1927law}, construct validity~\citep{cronbach1955construct,messick1995validity}, and cross-model agreement inspired by generalizability theory~\citep{brennan1992generalizability}. Unlike prior work that evaluates the reliability of individual judgments, we evaluate the reliability of automated evaluation \textit{systems}, including criteria generation, code synthesis, and plan-code alignment.
\section{Conclusion}
\label{sec:conclusion}
As AI agents move from research prototypes to production systems, the gap between agent capabilities and evaluation capabilities widens. Developers can build sophisticated multi-step agents, but evaluating them remains largely manual, requiring domain expertise to design metrics, engineering effort to instrument code, and tedious inspection to diagnose failures.

We introduced \textsc{EvalAgent} to close this gap. Its core insight is that general-purpose coding agents already have the capabilities for evaluation but lack evaluation-specific knowledge. \textsc{EvalAgent} encodes this missing knowledge as \textit{evaluation skills}---procedural instructions, reusable code and templates, and dynamically retrieved API documentation---that compose into a trace-based pipeline operating on runtime behavior (tool calls, error recovery, decision sequences) rather than source code alone. Our experiments across 20 diverse agents demonstrate that this approach achieves 84--100\% win-tie rates over baselines while maintaining 62.5--65.0\% Eval@1, with 79.5\% human expert preference.

Three practical takeaways emerge for practitioners building agent evaluation: (1)~incorporate execution traces whenever available---the 73--77\% quality improvement over static analysis is substantial; (2)~invest in evaluation skills rather than relying on general-purpose code generation---structured guidance prevents the metric proliferation and plan-code drift that plague unconstrained approaches; and (3)~ensure access to up-to-date API documentation, as the 45 percentage point Eval@1 gain from dynamic retrieval demonstrates this is not optional.

Beyond the system itself, we contribute AgentEvalBench, the first benchmark for agent evaluation automation, and a meta-evaluation framework validated by human experts. We hope these resources enable the research community to make progress on a problem that will only grow more important as agents become more capable and more widely deployed.

\paragraph{Limitations.}
(i) Our benchmark covers 20 agents but may not represent all agent types (\eg, embodied agents, multimodal agents). (ii) All experiments use the Claude model family; cross-model validation is needed to establish generalizability. (iii) Code executability of 62.5--65.0\% means approximately one-third of generated evaluations require manual debugging. (iv) Meta-evaluation itself is inherently challenging and subjective: assessing whether an evaluation is ``good'' requires nuanced judgment about what matters for a given agent, making this a fundamentally difficult problem even for human experts.

\paragraph{Future directions.}
Key directions include: (i) extending \textsc{EvalAgent} to support closed-loop iteration where evaluation feedback directly informs agent improvement; (ii) developing domain-specific evaluation skill libraries for different agent types (\eg, dialogue systems, embodied agents, coding assistants); (iii) advancing test data generation beyond simple synthesis to more sophisticated scenario construction; and (iv) establishing more generalizable trace format standardization. On the benchmark and methodology side, (v) expanding AgentEvalBench with additional agent types and (vi) developing more nuanced meta-evaluation methods remain important open challenges.


\newpage
\bibliography{main}

\newpage

\appendix

\section*{\LARGE Appendix}

\section{\textsc{EvalAgent} Pipeline Details}
\label{app:pipeline_details}

This section provides detailed descriptions of each phase in the \textsc{EvalAgent} pipeline, supplementing the overview in Section~\ref{sec:method}.

\subsection{Phase 1: Evaluation Planning}
\label{app:planning}

The evaluation plan is documented in a standardized Markdown format and includes:
\begin{itemize}
    \item \textbf{Agent Specification}: Description of the agent's purpose, capabilities, architecture, and technology stack.
    \item \textbf{Evaluation Objectives}: Clear, measurable goals aligned with user-specified requirements.
    \item \textbf{Test Scenarios}: High-level descriptions of interaction scenarios to be evaluated.
    \item \textbf{Evaluation Metrics}: 1--3 essential metrics with clear success criteria, expanding as needed.
    \item \textbf{Technology Stack}: Technical components for the evaluation (tracing libraries, evaluation frameworks, LLM providers, data formats).
\end{itemize}

\subsection{Phase 2: Test Case Generation}
\label{app:test-generation}

Test cases are stored in JSON Lines (JSONL) format. Figure~\ref{fig:test_case_example} shows an example structure. This phase is substitutable: alternative data generation approaches (\eg, \href{https://github.com/strands-agents/evals/tree/main/src/strands_evals/generators}{Strands Evals generators}) can be used as long as they emit test cases in the expected schema.

\begin{figure}[htbp]
\begin{minipage}[t]{\textwidth}
\captionsetup[algorithm]{name=Code}
\begin{algorithm}[H]
\caption{Example test case in JSONL format}
\label{fig:test_case_example}
\begin{lstlisting}[language=python]
'''
Each test case includes an identifier, scenario category, input query,
description, and expected behavior for evaluation.
'''
{
  "test_id": "general_info_001",
  "scenario": "General Information Queries",
  "query": "What are the latest developments in AI?",
  "description": "Tests agent's ability to search and synthesize",
  "expected_behavior": "Agent should perform web search and provide
                        a factual answer"
}
\end{lstlisting}
\end{algorithm}
\end{minipage}
\end{figure}

\subsection{Phase 3: Agent Instrumentation}
\label{app:instrumentation}

Figure~\ref{fig:tracing_code} shows an example using Traceloop~\citep{traceloop} for OTEL-compatible tracing. This lightweight instrumentation adds minimal overhead to the original codebase.

\begin{figure}[htbp]
\begin{minipage}[t]{\textwidth}
\captionsetup[algorithm]{name=Code}
\begin{algorithm}[H]
\caption{Example instrumentation code using Traceloop}
\label{fig:tracing_code}
\begin{lstlisting}[language=python]
'''
This minimal initialization enables OpenTelemetry-compatible tracing with
traces exported to a local collector.
'''
from traceloop.sdk import Traceloop

Traceloop.init(
    app_name="{agent-name}",
    disable_batch=True,
    api_endpoint="http://localhost:4318"
)
\end{lstlisting}
\end{algorithm}
\end{minipage}
\end{figure}

\subsection{Phase 4: Trace Collection and Processing}
\label{app:trace-collection}

\paragraph{Trace format.} \textsc{EvalAgent} expects traces in OpenTelemetry-compatible JSON format. Each trace consists of spans with timestamps, event types, and structured payloads.

\paragraph{Trace processing.} A trace processor converts OTEL JSONL into compact, per-trace JSON artifacts by filtering to agent-relevant spans (\eg, those carrying \texttt{traceloop.*}, \texttt{gen\_ai.*}, or \texttt{http.*} attributes) and extracting a minimal set of fields such as logical operation names, inputs/outputs, prompts/completions, tool metadata, and timing.

\subsection{Phase 5: Evaluation Code Generation}
\label{app:evaluation-codegen}

The system generates three main components:
\begin{enumerate}
    \item \textbf{Metric Implementations}: Python classes implementing specific evaluation logic, including both deterministic checks (\eg, tool usage patterns) and LLM-based assessments (\eg, response quality, task completion).
    \item \textbf{Evaluation Orchestrator}: Coordinates the evaluation pipeline: loading traces, applying metrics to each trace, and aggregating results.
    \item \textbf{Result Storage}: Persists evaluation outcomes in JSON format for analysis and reporting.
\end{enumerate}

\subsection{Phase 6: Reporting}
\label{app:reporting}

The evaluation report includes:
\begin{itemize}
    \item \textbf{Executive Summary}: High-level results including test scale, success rate, key strengths, and critical issues.
    \item \textbf{Results Analysis}: Detailed breakdown of metric performance across test scenarios.
    \item \textbf{Failure Analysis}: Systematic analysis with root cause identification and specific evidence.
    \item \textbf{Actionable Recommendations}: Prioritized, evidence-based suggestions for improving the target agent.
\end{itemize}

\section{AgentEvalBench Dataset Details}
\label{sec:agent-dataset-details}
This section provides detailed descriptions of 20 agents in the AgentEvalBench dataset.

\subsection{Simple Complexity Agents}

\paragraph{AI Tic-Tac-Toe Agent.}
\label{subsec:ai-tic-tac-toe}
\textit{Domain:} Game Playing.
\textit{Framework:} Agno.
\textit{Description:} A game-playing agent that implements the classic Tic-Tac-Toe game using a multi-agent architecture. The system demonstrates turn-based adversarial gameplay between two AI players.
\textit{Multi-Agent:} two agents (\texttt{player\_x} and \texttt{player\_o}) alternate turns in adversarial gameplay.

\paragraph{Airbnb MCP Agent.}
\label{subsec:airbnb-mcp}
\textit{Domain:} Travel Assistant.
\textit{Framework:} MCP/Bedrock.
\textit{Description:} A travel assistant agent that helps users search for and explore Airbnb listings. The agent leverages the Model Context Protocol (MCP) for tool integration with AWS Bedrock as the underlying LLM provider.
\textit{Tools:} 1 MCP tool for Airbnb search functionality.

\paragraph{Browser MCP Agent.}
\label{subsec:browser-mcp}
\textit{Domain:} Web Browsing.
\textit{Framework:} MCP/Bedrock.
\textit{Description:} A web browsing automation agent that can navigate websites, interact with page elements, and capture screenshots. The agent uses MCP tools to control browser actions programmatically.
\textit{Tools:} 4 MCP tools: \texttt{browser\_navigate} (navigate to URLs), \texttt{browser\_click} (click on page elements), \texttt{browser\_evaluate} (execute JavaScript), \texttt{browser\_take\_screenshot} (capture page screenshots).

\paragraph{Career Assist.}
\label{subsec:career-assist}
\textit{Domain:} Career Planning.
\textit{Framework:} LangChain/LangGraph.
\textit{Description:} A career planning assistant that helps users with various career-related tasks including job searching, resume creation, interview preparation, and finding learning resources.
\textit{Tools:} 5 LangGraph handlers: \texttt{categorize}, \texttt{handle\_learning\_resource}, \texttt{handle\_resume\_making}, \texttt{handle\_interview\_preparation}, \texttt{handle\_job\_search}.

\paragraph{Chat with PDF Agent.}
\label{subsec:chat-pdf}
\textit{Domain:} Document QA.
\textit{Framework:} Embedchain.
\textit{Description:} A document question-answering agent that allows users to upload PDF documents and ask questions about their content. The agent uses vector embeddings for semantic search over document content.
\textit{Tools:} 2 functions: \texttt{load\_pdf} (load and process PDF documents), \texttt{ask\_question} (query the document content).
\textit{Memory:} ChromaDB vector database for document embeddings and retrieval.

\paragraph{Chat with Research Papers.}
\label{subsec:chat-research}
\textit{Domain:} Research Assistant.
\textit{Framework:} Agno.
\textit{Description:} A research assistant agent that helps users find and discuss academic papers from arXiv. The agent can search for papers based on topics and answer questions about research content.
\textit{Tools:} \texttt{ArxivTools} (search and retrieve papers from arXiv).

\paragraph{Local News Agent.}
\label{subsec:local-news}
\textit{Domain:} News Aggregation.
\textit{Framework:} Custom.
\textit{Description:} A news aggregation agent that searches for, synthesizes, and summarizes local news. The agent employs a multi-agent pipeline architecture for comprehensive news processing.
\textit{Tools:} \texttt{DuckDuckGoSearchResults} (web search for news articles).
\textit{Multi-Agent:} three specialized agents: \texttt{search\_agent} (finds relevant news articles), \texttt{synthesis\_agent} (combines information from multiple sources), \texttt{summary\_agent} (creates concise summaries).

\paragraph{MiniSWE Agent.}
\label{subsec:minisweagent}
\textit{Domain:} Software Engineering.
\textit{Framework:} Custom.
\textit{Description:} A minimal software engineering agent designed for code understanding and modification tasks. The agent represents a lightweight approach to automated software engineering.

\paragraph{Simple QA Search.}
\label{subsec:simple-qa}
\textit{Domain:} Question Answering.
\textit{Framework:} LangGraph.
\textit{Description:} A straightforward question-answering agent that combines web search with LLM reasoning to answer user queries. The agent demonstrates basic retrieval-augmented generation patterns.
\textit{Tools:} 1 search tool for web queries.

\paragraph{Single Agent Trip Crew.}
\label{subsec:trip-crew}
\textit{Domain:} Trip Planning.
\textit{Framework:} CrewAI.
\textit{Description:} A trip planning agent built on the CrewAI framework that helps users plan travel itineraries by searching for destinations and scraping relevant travel information.
\textit{Tools:} 2 tools: \texttt{SerperDevTool} (search engine integration), \texttt{ScrapeWebsiteTool} (web content extraction).

\subsection{Medium Complexity Agents}

\paragraph{City Agent.}
\label{subsec:city-agent}
\textit{Domain:} Information Retrieval.
\textit{Framework:} LangGraph.
\textit{Description:} An information retrieval agent that provides city-related information including events, restaurant recommendations, search results, and weather data. The agent integrates multiple data sources for comprehensive city information.
\textit{Tools:} 4 tool classes: \texttt{EventsDatabaseTool} (query local events), \texttt{RestaurantRecommendationTool} (find restaurants), \texttt{SearchTool} (general information search), \texttt{WeatherTool} (weather forecasts).

\paragraph{CleverChatty Agent.}
\label{subsec:cleverchatty}
\textit{Domain:} Conversational AI.
\textit{Framework:} MCP.
\textit{Description:} A conversational AI agent with persistent memory capabilities. The agent can remember information from past conversations and recall it in future interactions, enabling more contextual and personalized conversations.
\textit{Tools:} 3 MCP memory tools: \texttt{remember} (store information for later recall), \texttt{recall} (retrieve previously stored information), \texttt{search\_in\_memory} (search through stored memories).
\textit{Memory:} \texttt{MCPMemoryClient} with \texttt{remember()} and \texttt{recall()} methods for persistent conversation memory.

\paragraph{Code Assistant.}
\label{subsec:code-assistant}
\textit{Domain:} Code Generation.
\textit{Framework:} Strands.
\textit{Description:} A code generation and review agent that can read projects, generate code, review existing code, write files, and execute code. The agent provides comprehensive software development assistance.
\textit{Tools:} 5 \texttt{@tool} decorated functions: \texttt{project\_reader} (read and analyze project structure), \texttt{code\_generator} (generate new code), \texttt{code\_reviewer} (review code quality), \texttt{code\_writer\_agent} (write code to files), \texttt{code\_execute} (execute code snippets).

\paragraph{Data Warehouse Optimizer.}
\label{subsec:data-warehouse}
\textit{Domain:} Database Optimization.
\textit{Framework:} Strands.
\textit{Description:} A database optimization agent that analyzes SQL queries, suggests optimizations, and validates query costs. The agent uses a multi-agent workflow with specialized agents for different optimization tasks.
\textit{Tools:} 4 \texttt{@tool} decorated functions: \texttt{get\_query\_execution\_plan} (analyze query execution), \texttt{suggest\_optimizations} (recommend improvements), \texttt{validate\_query\_cost} (estimate query costs), \texttt{calculator} (perform calculations).
\textit{Multi-Agent:} three agents in a LangGraph workflow: \texttt{analyzer\_agent} (analyzes query patterns), \texttt{rewriter\_agent} (rewrites queries for optimization), \texttt{validator\_agent} (validates optimized queries).

\paragraph{LlamaIndex Complex Agent.}
\label{subsec:llamaindex-complex}
\textit{Domain:} Financial Analysis.
\textit{Framework:} LlamaIndex.
\textit{Description:} A financial analysis agent that uses multiple document indices for retrieval-augmented generation. The agent can analyze financial documents, calculate metrics, and compare companies.
\textit{Tools:} 5 tools: 3 \texttt{QueryEngineTools} (query different document indices), \texttt{calculate\_financial\_metric} (compute financial metrics), \texttt{compare\_companies} (compare company data).
\textit{Memory:} \texttt{VectorStoreIndex} multi-index RAG system with chat memory buffer for maintaining conversation context.

\paragraph{Medical Document Processor.}
\label{subsec:medical-doc}
\textit{Domain:} Medical NLP.
\textit{Framework:} Strands.
\textit{Description:} A medical document processing agent that extracts clinical information and links it to standard medical terminologies (ICD, RxNorm, SNOMED). The agent is designed for healthcare NLP applications.
\textit{Tools:} 8 tools: \texttt{file\_read}, \texttt{process\_document}, \texttt{get\_icd}, \texttt{get\_rx}, \texttt{get\_snomed}, \texttt{link\_icd}, \texttt{link\_rx}, \texttt{link\_snomed}.

\subsection{Hard Complexity Agents}

\paragraph{Adala Agent.}
\label{subsec:adala}
\textit{Domain:} Data Labeling.
\textit{Framework:} Custom-Adala.
\textit{Description:} A sophisticated data labeling agent built on the Adala framework. The agent supports multiple labeling tasks and includes a learning system that improves labeling quality over time through memory-based feedback.
\textit{Tools:} 9 skills: \texttt{classification}, \texttt{summarization}, \texttt{qa}, \texttt{translation}, \texttt{entity\_extraction}, \texttt{text\_generation}, \texttt{rag}, \texttt{ontology\_creator}, \texttt{multilabel\_classification}.
\textit{Memory:} \texttt{VectorDBMemory} class with \texttt{remember()}, \texttt{retrieve()}, \texttt{FileMemory}, and \texttt{LearningSystem} for adaptive learning from feedback.

\paragraph{Agent4Rec Agent.}
\label{subsec:agent4rec}
\textit{Domain:} Recommendation.
\textit{Framework:} LangChain.
\textit{Description:} A recommendation system agent that simulates user interactions to generate personalized recommendations. The agent uses memory with time and importance weighting for more accurate preference modeling.
\textit{Tools:} 3 tools: Arena simulation (simulate recommendation scenarios), Avatar (user preference modeling), Agent ask method (query the recommendation system).
\textit{Memory:} \texttt{abstract\_memory.py} with \texttt{add\_memory()}, \texttt{time\_weighting()}, \texttt{importance\_weighting()}, and \texttt{reflect()} methods.

\paragraph{Hilton Search Agent.}
\label{subsec:hilton-search}
\textit{Domain:} Hotel Search.
\textit{Framework:} LangChain/LangGraph.
\textit{Description:} A hotel search agent specialized for Hilton properties. The agent can search hotels, calculate costs, and help users utilize loyalty points for bookings. It maintains session-based memory for personalized search experiences.
\textit{Tools:} 3 \texttt{@tool} decorated functions: \texttt{search\_hotels} (search Hilton properties), \texttt{get\_hotel\_cost} (calculate booking costs), \texttt{utilize\_points} (apply loyalty points).
\textit{Memory:} ChromaDB vectorstore with session management for maintaining search context and user preferences across interactions.

\paragraph{Network Switch Operator Agent.}
\label{subsec:network-switch}
\textit{Domain:} Network Management.
\textit{Framework:} Custom (Bedrock).
\textit{Description:} A network management agent that monitors and operates network switches. The agent can retrieve device information, perform health checks, analyze network topology, and execute device reboots.
\textit{Tools:} 4 tool methods: \texttt{get\_device\_summary} (retrieve device information), \texttt{reboot\_device} (restart network devices), \texttt{get\_network\_topology} (map network structure), \texttt{check\_device\_health} (monitor device status).

\subsection{Scenario Examples}
\label{sec:example-scenarios}

Test scenarios use JSON format with agent-specific schemas. Figure~\ref{fig:scenario-json} shows examples.

\begin{figure}[htbp]
\begin{minipage}[t]{\textwidth}
\captionsetup[algorithm]{name=Code}
\begin{algorithm}[H]
\caption{Example JSON scenarios from AgentEvalBench}
\label{fig:scenario-json}
    \begin{lstlisting}[language=python]
# Simple query format
{"query": "Find Hilton hotels in Miami with pool amenities"}

{"query": "Create a Python script that calculates simple 
  and compound interest"}

# Structured format (classification agents)
{
  "data": [{"text": "Great product!"}, {"text": "Terrible"}],
  "skill": "classification",
  "parameters": {"labels": ["positive", "negative"]}
}

# Domain-specific format (trip planning)
{
  "trip_request": {
    "origin": "Los Angeles, LAX",
    "destination": "Bali, Indonesia",
    "duration": "10 days"
  }
}
    \end{lstlisting}
\end{algorithm}
\end{minipage}
\end{figure}

\subsection{Evaluation Requirements Examples}
\label{sec:eval-req-examples}

Each agent has generic and specific evaluation requirements in YAML format. Figure~\ref{fig:eval-req} shows examples.

\begin{figure}[htbp]
\begin{minipage}[t]{\textwidth}
\captionsetup[algorithm]{name=Code}
\begin{algorithm}[H]
\caption{Evaluation requirements examples}
\label{fig:eval-req}
    \begin{lstlisting}[language=python]
'''
Generic requirements test whether methods can infer metrics
without guidance; specific requirements provide domain-specific
evaluation criteria.
'''

# Generic requirement
"Given the agent codebase at ./src and execution
traces at ./data/trace, create an evaluation plan"

# Specific requirements

# medical_document_processor
"...create an evaluation plan to evaluate
correctness of extracted medical entities
and diagnoses"

# hilton_search_agent
"...create an evaluation plan to evaluate
how well hotel results match location, dates,
and amenity requirements"

# code_assistant
"...create an evaluation plan to evaluate
correctness, efficiency, and best practices
in generated code"
    \end{lstlisting}
\end{algorithm}
\end{minipage}
\end{figure}

\section{Meta-Evaluation Dimension Rubrics}
\label{app:rubrics}

We provide detailed comparative rubrics for each evaluation dimension used in meta-evaluation. The meta-evaluation is purely comparative with no numerical scores: for each dimension, the meta-evaluator determines whether A Wins, B Wins, or Tie by directly comparing the two approaches.

\paragraph{Winner determination.} For each dimension, the overall winner is calculated by awarding points: the winner of a dimension receives the full weight as points, while a tie means both approaches receive half the weight. The approach with the higher total points wins overall.

\subsection{User Requirement Fulfillment (Weight: 15\%)}
\label{app:rubric_user_req}

This dimension assesses which evaluation more effectively addresses explicit user requirements.

\begin{table}[h]
\centering
\footnotesize
\setlength{\tabcolsep}{4pt}
\renewcommand{\arraystretch}{1.2}
\begin{tabularx}{\linewidth}{l X}
\toprule
Outcome & Comparative Criteria \\
\graycmidrule{1-2}
A Wins & Fulfills requirements opponent misses/addresses less directly; provides more focused implementation. \\
B Wins & Same as A Wins with A$\leftrightarrow$B swapped. \\
Tie & Both fulfill requirements equally, OR requirement is generic (\eg, ``create an evaluation plan'') with nothing explicit to compare. \\
\bottomrule
\end{tabularx}
\captionsetup{justification=centering}
\caption{\textbf{Rubric for User Requirement Fulfillment.} Focused implementation beats exhaustive additions.}
\label{tab:rubric_user_req}
\end{table}

\subsection{Metric Relevance (Weight: 30\%)}
\label{app:rubric_metric}

This dimension evaluates which codebase's metrics have a better signal-to-noise ratio, receiving the highest weight due to its importance in evaluation quality. Signal refers to metrics that evaluate what matters for the evaluation goal; noise refers to metrics that do not add meaningful value (trivial, redundant, or distracting).

\begin{table}[h]
\centering
\footnotesize
\setlength{\tabcolsep}{4pt}
\renewcommand{\arraystretch}{1.2}
\begin{tabularx}{\linewidth}{l X}
\toprule
Outcome & Comparative Criteria \\
\graycmidrule{1-2}
A Wins & Better S/N ratio—covers essentials with less noise; more relevant metrics, less redundancy, clearer names. \\
B Wins & Same as A Wins with A$\leftrightarrow$B swapped. \\
Tie & Same S/N ratio, comparable redundancy/counts, or trade-offs balance out; similar metric clarity. \\
\bottomrule
\end{tabularx}
\captionsetup{justification=centering}
\caption{\textbf{Rubric for Metric Relevance.} More metrics $\neq$ better signal. Tie-breaker: metric name clarity.}
\label{tab:rubric_metric}
\end{table}

\subsection{Code Quality \& Complexity (Weight: 25\%)}
\label{app:rubric_code}

This dimension evaluates which codebase has correct metric logic, cleaner implementation, and better organization. Win criteria are applied in priority order: (1) metric logic correctness, (2) code organization, (3) clean implementation, (4) code readability.

\begin{table}[h]
\centering
\footnotesize
\setlength{\tabcolsep}{4pt}
\renewcommand{\arraystretch}{1.2}
\begin{tabularx}{\linewidth}{l X}
\toprule
Outcome & Comparative Criteria \\
\graycmidrule{1-2}
A Wins & Correct metric logic (opponent has $\geq$1 flaw); simpler organization (fewer files, shorter code) with equivalent functionality; no unused imports/over-engineering (opponent has some); clearer names. \\
B Wins & Same as A Wins with A$\leftrightarrow$B swapped. \\
Tie & Both have correct metric logic, similar file count/length, clean implementations, similar readability. \\
\bottomrule
\end{tabularx}
\captionsetup{justification=centering}
\caption{\textbf{Rubric for Code Quality \& Complexity.}}
\label{tab:rubric_code}
\end{table}

\subsection{Plan Quality (Weight: 15\%)}
\label{app:rubric_plan}

This dimension assesses which evaluation plan is more coherent, complete, actionable, and efficiently documented. Win criteria are applied in priority order: (1) plan completeness, (2) definition completeness, (3) conciseness, (4) ease of understanding.

\begin{table}[h]
\centering
\footnotesize
\setlength{\tabcolsep}{4pt}
\renewcommand{\arraystretch}{1.2}
\begin{tabularx}{\linewidth}{l X}
\toprule
Outcome & Comparative Criteria \\
\graycmidrule{1-2}
A Wins & More complete structure (goals, metrics, methodology) while opponent misses essentials; equivalent completeness with $\geq$50 fewer lines; more readable. \\
B Wins & Same as A Wins with A$\leftrightarrow$B swapped. \\
Tie & Both have all required components, complete definitions, similar length and readability. \\
\bottomrule
\end{tabularx}
\captionsetup{justification=centering}
\caption{\textbf{Rubric for Plan Quality.} Plans $>$1000 lines need justification; $>$1500 lines likely loses to shorter equivalent.}
\label{tab:rubric_plan}
\end{table}

\subsection{Plan-Code Alignment (Weight: 15\%)}
\label{app:rubric_alignment}

This dimension verifies which codebase's implementation more faithfully follows its plan. Win criteria are applied in priority order: (1) metric alignment, (2) implementation mismatches, (3) missing implementations, (4) unplanned features.

\begin{table}[h]
\centering
\footnotesize
\setlength{\tabcolsep}{4pt}
\renewcommand{\arraystretch}{1.2}
\begin{tabularx}{\linewidth}{l X}
\toprule
Outcome & Comparative Criteria \\
\graycmidrule{1-2}
A Wins & All metrics aligned (opponent has $\geq$1 misaligned); zero implementation mismatches (opponent has $\geq$1); all planned features implemented (opponent missing $\geq$1); zero unplanned features (opponent has $\geq$1). \\
B Wins & Same as A Wins with A$\leftrightarrow$B swapped. \\
Tie & Both have all metrics aligned, no mismatches, missing/unplanned features within 1 each. \\
\bottomrule
\end{tabularx}
\captionsetup{justification=centering}
\caption{\textbf{Rubric for Plan-Code Alignment.}}
\label{tab:rubric_alignment}
\end{table}

\section{Prompt Specifications}
\label{app:prompts}

This section provides the key prompts used in \textsc{EvalAgent}. Due to space constraints, we present structured summaries with key components highlighted. Full prompt templates are available at \url{https://github.com/awslabs/Agent-EvalKit}.

\subsection{Baseline Approaches}
\label{app:baseline_approaches}

Table~\ref{tab:baseline-approaches} summarizes the baseline approaches compared against \textsc{EvalAgent}.

\begin{table}[htbp]
\centering
\footnotesize
\begin{tabular}{@{}lccp{4cm}@{}}
\toprule
\textbf{Approach} & \textbf{Stages} & \textbf{Tool Access} & \textbf{Context Delivery} \\
\graycmidrule{1-4}
LLM-SingleTurn (B1) & 1 & None & Embedded in prompt \\
Agent-SourceCode (B2) & 1 & Full & Filesystem (no traces) \\
Agent-OneStage (B3) & 1 & Full & Filesystem \\
Agent-TwoStage (B4) & 2 & Full & Filesystem \\
\textsc{EvalAgent} (Ours) & 2 & Full & Filesystem + templates \\
\bottomrule
\end{tabular}
\captionsetup{justification=centering}
\caption{\textbf{Baseline approaches for evaluation code generation.}}
\label{tab:baseline-approaches}
\end{table}

\begin{tcolorbox}[colback=gray!5,colframe=gray!75,title=B1: LLM Single-Turn Baseline]
\small
\textbf{Context Delivery:} All context embedded directly in prompt (no tool access)

\textbf{Prompt Structure:}
\begin{verbatim}
# Agent Source Code:
## File: {path}
{file_contents for all source files}

# EXECUTION TRACES
{trace contents}

# EVALUATION TASK
{evaluation_requirements}

## Output Format (STRICT)
### File: <filename>.py
\end{verbatim}

\textit{Note: Pure text completion with strict output format. No iterative exploration.}
\end{tcolorbox}

\begin{tcolorbox}[colback=gray!5,colframe=gray!75,title=B2: Agent Source-Code Baseline]
\small
\textbf{Context:} Agent source code via filesystem, \textbf{no execution traces}

\textbf{Prompt:}
\begin{verbatim}
Given the agent codebase, directly implement 
evaluation code in Python.
\end{verbatim}

\textbf{Tool Access:} Full agent tools (file operations, search, code editing, shell commands)
\end{tcolorbox}

\begin{tcolorbox}[colback=gray!5,colframe=gray!75,title=B3: Agent One-Stage Baseline]
\small
\textbf{Context:} Agent source code + execution traces via filesystem

\textbf{Prompt:}
\begin{verbatim}
Given the agent codebase and execution traces, 
directly implement evaluation code in Python.
\end{verbatim}

\textbf{Tool Access:} Full agent tools (file operations, search, code editing, shell commands)

\textit{Note: Single-stage generation without intermediate planning.}
\end{tcolorbox}

\begin{tcolorbox}[colback=gray!5,colframe=gray!75,title=B4: Agent Two-Stage Baseline]
\small
\textbf{Context:} Agent source code + execution traces via filesystem

\textbf{Stage 1 (Plan) Prompt:}
\begin{verbatim}
Given the agent codebase and execution traces, 
create an evaluation plan.
\end{verbatim}

\textbf{Stage 2 (Code) Prompt:}
\begin{verbatim}
Given the evaluation plan, implement evaluation code. 
\end{verbatim}

\textbf{Tool Access:} Full agent tools for both stages

\textit{Note: Two-stage without structured templates or reference implementations.}
\end{tcolorbox}

\subsection{\textsc{EvalAgent} Prompts}
\label{app:prompt_planning}

\begin{tcolorbox}[colback=gray!5,colframe=gray!75,title=Planning Prompt]
\small
\textbf{Role:} Evaluation Strategy Designer

\textbf{Input:} Agent source code, execution traces (OTEL format), user requirements

\textbf{Workflow:}
\begin{enumerate}[noitemsep,topsep=0pt,leftmargin=*]
    \item Analyze agent architecture, capabilities, and behavior patterns
    \item Review execution traces to understand runtime behavior
    \item Identify 2--4 key metrics capturing core agent behaviors
    \item Design test scenarios exercising critical functionality
\end{enumerate}

Focus on actionable content. Each metric must measure a distinct behavioral aspect.

\textbf{Output:} Structured evaluation plan with agent analysis, evaluation goals, metrics with scoring rubrics, test scenarios, and implementation notes.

\end{tcolorbox}

\begin{tcolorbox}[colback=gray!5,colframe=gray!75,title=Code Generation Prompt]
\small
\textbf{Role:} Evaluation Code Implementer

\textbf{Input:} Evaluation plan, OTEL traces

\textbf{Workflow:}
\begin{enumerate}[noitemsep,topsep=0pt,leftmargin=*]
    \item Implement metrics from evaluation plan
    \item Create metric classes with extraction functions for raw OTEL data
    \item Build main entry point for evaluation pipeline
    \item Code review and fix critical issues
    \item Update dependencies
\end{enumerate}

\textbf{Key Principles:} Create minimal working version first. Validate library APIs before implementation. Avoid over-engineering; follow plan exactly.

\textbf{Code Quality:} Target 200--400 LOC, 2--4 files, no unnecessary abstractions.

\textbf{Output:} Metrics implementation, evaluation runner, requirements file
\end{tcolorbox}

\subsection{Evaluation Skills Examples}
\label{app:guidance_examples}

\textsc{EvalAgent}'s evaluation skills comprise procedural instructions, reusable code and templates, and dynamic resources. Together, they standardize the end-to-end workflow, reduce run-to-run variance, and improve Eval@1.

\begin{tcolorbox}[colback=gray!5,colframe=gray!75,title=Evaluation Code Generation -- Context7 Validation Requirement]
\small
\begin{verbatim}
### Context7 Validation

**MANDATORY**: Before implementing functionality from any publicly
available evaluation library, validate its current API usage with
Context7 MCP:

- "What's the current DeepEval LLMTestCase constructor signature?"
- "Show me the latest DeepEval BaseMetric implementation patterns"
- "What are the current built-in metrics available in DeepEval?"

This ensures production-ready code that follows current best practices
and avoids deprecated patterns.
\end{verbatim}
\end{tcolorbox}

\paragraph{Reusable recipes: plan/report templates.}
Templates provide structured scaffolding for evaluation artifacts. The plan template defines sections for agent analysis, metrics specification, and progress tracking. The report template provides structured scaffolding for evaluation reports, ensuring consistent documentation of results, analysis, and recommendations.

\begin{tcolorbox}[colback=gray!5,colframe=gray!75,title=Plan Template -- Evaluation Metrics Section]
\small
\begin{verbatim}
## 3. Evaluation Metrics
<!-- ACTION REQUIRED: If no specific user requirements are provided,
use a minimal number of metrics (1-2 metrics) focusing on the most
critical aspects of agent performance. -->

### [Metric Name 1]
- **Evaluation Area:** [Final response quality/tool call accuracy/...]
- **Description:** [What is measured and why]
- **Method:** [Code-based | LLM-as-Judge]
\end{verbatim}
\end{tcolorbox}

\begin{tcolorbox}[colback=gray!5,colframe=gray!75,title=Excerpt: \texttt{eval-report-template.md} -- Executive Summary Section]
\small
\begin{verbatim}
# Agent Evaluation Report for [AGENT NAME]

## Executive Summary
- **Test Scale**: [N] test cases
- **Success Rate**: [XX.X%]
- **Status**: [Excellent/Good/Poor]
- **Strengths**: [Specific capability] [Performance highlight]
- **Critical Issues**: [Blocking issue + impact] [Bottleneck]
- **Action Priority**: [Critical fixes] [Improvements]

## Agent Failure Analysis
### Issue 1 - [Priority Level]
- **Issue**: [Clear problem statement with evaluation metrics]
- **Root Cause**: [Technical analysis — path/to/file.py:START-END]
- **Priority Fixes**:
  - P1 — [Fix name]: [Solution] → Expected gain: [Metric +X]
\end{verbatim}
\end{tcolorbox}

\paragraph{Reusable recipes: code patterns.}
Code patterns provide implementation scaffolding for trace extraction and metrics. The patterns demonstrate OTEL JSONL parsing and DeepEval integration.

\begin{tcolorbox}[colback=gray!5,colframe=gray!75,title=Trace Extraction Pattern]
\small
\begin{verbatim}
def get_span_attribute(span: Dict, key: str) -> Optional[str]:
    """Extract attribute value from OTEL span attributes array."""
    for attr in span.get("attributes", []):
        if attr.get("key") == key:
            value = attr.get("value", {})
            return (value.get("stringValue") or value.get("intValue")
                    or value.get("boolValue") or value.get("doubleValue"))
    return None

def load_trace_to_test_case(file_path: str) -> LLMTestCase:
    """Convert OTEL JSONL trace file to DeepEval test case."""
    spans = load_spans_from_jsonl(file_path)
    return LLMTestCase(
        input=extract_user_input(spans),
        actual_output=extract_final_response(spans)
    )
\end{verbatim}
\end{tcolorbox}

\begin{tcolorbox}[colback=gray!5,colframe=gray!75,title=Custom Metric Pattern]
\small
\begin{verbatim}
class TaskCompletionMetric(BaseMetric):
    """Custom metric using G-Eval for task completion assessment."""
    def __init__(self, threshold: float = 0.7):
        self.threshold = threshold
        self.geval = GEval(
            name="Task Completion",
            criteria="Evaluate whether the agent successfully completed
                      the requested task based on input and output.",
            evaluation_params=[LLMTestCaseParams.INPUT,
                               LLMTestCaseParams.ACTUAL_OUTPUT],
            model="claude-3-5-sonnet-20241022"
        )
    
    def measure(self, test_case: LLMTestCase) -> float:
        self.geval.measure(test_case)
        self.score = self.geval.score
        return self.score
\end{verbatim}
\end{tcolorbox}

\subsection{Meta-Evaluator Comparison Prompt}
\label{app:prompt_metaeval}

\begin{tcolorbox}[colback=gray!5,colframe=gray!75,title=Meta-Evaluator Prompt]
\small
\textbf{Role:} Comparative Meta-Evaluator

\textbf{Task:} Compare evaluation system A \vs B to determine which is superior

\textbf{Evaluation Dimensions (weighted):}
\begin{enumerate}[noitemsep,topsep=0pt,leftmargin=*]
    \item User Requirement Fulfillment (15\%) -- explicit requirement coverage
    \item Metric Relevance (30\%) -- signal-to-noise ratio of metrics
    \item Code Quality \& Complexity (25\%) -- correctness, organization
    \item Plan Quality (15\%) -- coherence, completeness, conciseness
    \item Plan-Code Alignment (15\%) -- faithfulness to plan
\end{enumerate}

\textbf{Anti-Length Bias:} Conciseness is a virtue. Quality over quantity (fewer focused metrics preferred). Practitioner perspective: ``Which would a developer maintain?'' Red flags: 20+ metrics, code 2$\times$ necessary length, plan $>$1500 lines.

\textbf{Workflow:} Navigate to each approach's directory $\rightarrow$ Read plans and code $\rightarrow$ Determine winner per dimension (A/B/Tie) $\rightarrow$ Calculate weighted score $\rightarrow$ Provide evidence-based justification

\textbf{Output:} Per-dimension winners with justification, points calculation (out of 1.00), overall winner, key differentiating factors
\end{tcolorbox}

\section{Human Annotation Study}
\label{app:annotation}

\subsection{Study Design}
\label{app:annotator_selection}
Three expert annotators ($\geq$3 years ML/AI engineering experience, agent development expertise, no prior involvement in \textsc{EvalAgent}) independently evaluated 40 agent-evaluation pairs across 5 dimensions (200 judgments total). Annotators were blind to system identity with randomized A/B assignment. Each completed a 2-hour training session with 5 practice annotations and a calibration round. Average annotation time was 68 minutes per case.

\subsection{Annotation Instructions}
\label{app:annotation_instructions}

The following instructions were provided to annotators:

\begin{tcolorbox}[colback=blue!5,colframe=blue!75,title=Human Annotation Instructions]
\small
\textbf{Task:} You will evaluate pairs of agent evaluation artifacts (plans and code) and determine which approach produced higher-quality evaluation using pairwise comparison.

\textbf{Materials Provided:}
\begin{itemize}
    \item Agent source code and execution traces
    \item Evaluation Plan A and Evaluation Code A
    \item Evaluation Plan B and Evaluation Code B
    \item Comparative dimension rubrics (see Appendix B)
\end{itemize}

\textbf{Procedure:}
\begin{enumerate}
    \item Review the agent under evaluation (15-20 minutes)
    \item Read Evaluation Plan A, then Evaluation Code A (20-30 minutes)
    \item Read Evaluation Plan B, then Evaluation Code B (20-30 minutes)
    \item For each of the 5 dimensions, determine the winner: \textbf{A Wins}, \textbf{B Wins}, or \textbf{Tie}
    \item Calculate overall winner based on weighted dimension outcomes
    \item Provide brief justification for each dimension winner
\end{enumerate}

\textbf{Important Guidelines:}
\begin{itemize}
    \item Focus on evaluation quality, not agent quality
    \item Apply comparative rubrics consistently across all annotations
    \item Declare a tie only when approaches are genuinely equivalent
    \item Take breaks between annotations to maintain focus
\end{itemize}
\end{tcolorbox}

\subsection{Inter-Annotator Agreement}
\label{app:iaa_details}

\begin{table}[htbp]
\centering
\footnotesize
\setlength{\tabcolsep}{4pt}
\renewcommand{\arraystretch}{1.15}
\begin{tabular}{l c c}
\toprule
Dimension & Fleiss' $\kappa$ & \makecell{Avg Pairwise\\Agreement} \\
\graycmidrule{1-3}
User Requirement Fulfillment & 0.937 & 96.7\% \\
Metric Relevance & 1.000 & 100.0\% \\
Code Quality \& Complexity & 0.948 & 98.3\% \\
Plan Quality & 0.885 & 96.7\% \\
Plan-Code Alignment & 0.700 & 95.0\% \\
\graycmidrule{1-3}
\textbf{Overall} & \textbf{0.923} & \textbf{97.3\%} \\
\bottomrule
\end{tabular}
\caption{\textbf{Inter-annotator agreement by dimension.} Fleiss' $\kappa$ across 3 annotators on 40 cases. All dimensions achieve almost perfect agreement ($\kappa > 0.80$) except Plan-Code Alignment ($\kappa = 0.700$, substantial), reflecting greater subjectivity in assessing implementation faithfulness.}
\label{tab:iaa_details}
\end{table}

Table~\ref{tab:iaa_details} reports per-dimension inter-annotator agreement. Metric Relevance achieves perfect agreement ($\kappa = 1.000$), while Plan-Code Alignment shows the lowest ($\kappa = 0.700$), reflecting the greater subjectivity in assessing implementation faithfulness. Majority vote was used for consensus; overall winner disagreement occurred in 8/40 cases (20\%).
%


\subsection{Meta-Evaluator \vs Human Expert Alignment}
\label{app:llmj_alignment_detail}

Table~\ref{tab:llmj_alignment_full} compares two meta-evaluator models against human majority vote on the same 40 pairs across 5 dimensions (200 judgments).

\begin{table}[htbp]
\centering
\footnotesize
\renewcommand{\arraystretch}{1.15}
\begin{tabular}{l c c}
\toprule
\textbf{Metric} & \textbf{Sonnet 4.5} & \textbf{Opus 4.5} \\
\graycmidrule{1-3}
\multicolumn{3}{l}{\textit{Case-level (40 pairs, weighted overall winner)}} \\
Overall winner match & 90.0\% (36/40) & 97.5\% (39/40) \\
\graycmidrule{1-3}
\multicolumn{3}{l}{\textit{Per-judgment (200 dimension-level decisions)}} \\
\quad Accuracy & 0.755 {\scriptsize [.695, .815]} & 0.750 {\scriptsize [.695, .815]} \\
\quad Gwet's AC1 & 0.696 {\scriptsize [.615, .774]} & 0.687 {\scriptsize [.607, .771]} \\
\quad Spearman $\rho$ & 0.624 & 0.617 \\
\quad Macro-F1 & 0.561 {\scriptsize [.477, .639]} & 0.550 {\scriptsize [.476, .623]} \\
\graycmidrule{1-3}
\multicolumn{3}{l}{\textit{Per-dimension agreement (\% matching human majority vote)}} \\
\quad User Requirement Fulfillment & 85.0\% (34/40) & 95.0\% (38/40) \\
\quad Metric Relevance & 85.0\% (34/40) & 97.5\% (39/40) \\
\quad Code Quality \& Complexity & 72.5\% (29/40) & 72.5\% (29/40) \\
\quad Plan Quality & 75.0\% (30/40) & 50.0\% (20/40) \\
\quad Plan-Code Alignment & 60.0\% (24/40) & 60.0\% (24/40) \\
\textbf{Dimension avg} & \textbf{75.5\%} (151/200) & \textbf{75.0\%} (150/200) \\
\bottomrule
\end{tabular}
\caption{\textbf{Meta-evaluator alignment with expert consensus} (\textsc{EvalAgent} \vs B4, $n{=}40$ cases). Case-level: weighted overall winner match. Per-judgment: agreement on 200 dimension-level decisions (Gwet's AC1 and Spearman $\rho$ are chance-corrected). Weakest per-dimension agreement is on PCA, consistent with the lower inter-annotator $\kappa$ for that dimension (Table~\ref{tab:iaa_details}).}
\label{tab:llmj_alignment_full}
\end{table}

\section{Detailed Efficiency Analysis}
\label{app:efficiency}

We provide detailed computational efficiency breakdowns for the evaluation approaches.

\subsection{Efficiency by Requirement Specificity}
\label{app:efficiency_specificity}

Table~\ref{tab:efficiency_specificity} presents the computational efficiency breakdown by requirement specificity. For agentic methods with traces (B3 and B4), specific requirements help constrain the search space, reducing exploration overhead. \textsc{EvalAgent} maintains stable efficiency across both conditions. 

Model choice affects efficiency patterns: Haiku executes faster but consumes more tokens, while Sonnet is slower but more token-efficient. \textsc{EvalAgent} maintains its efficiency advantages across both models.

\begin{table}[h]
\centering
\footnotesize
\setlength{\tabcolsep}{4pt}
\renewcommand{\arraystretch}{1.2}
\begin{tabular}{c c c c c c c c}
\toprule
\multirow{2}{*}{LLM} & \multirow{2}{*}{Approach} & \multicolumn{3}{c}{\textbf{Generic}} & \multicolumn{3}{c}{\textbf{Specific}} \\
\cmidrule(lr){3-5} \cmidrule(lr){6-8}
& & Time (min) & Tokens (K) & Cost (\$) & Time (min) & Tokens (K) & Cost (\$) \\
\midrule
\multirow{5}{*}{Haiku 4.5} & B1 & 0.67 & 50.27 & 0.25 & 0.69 & 50.26 & 0.25 \\
& B2 & 3.86 & 1858.71 & 1.57 & 5.23 & 2533.60 & 2.10 \\
& B3 & 5.61 & 3473.38 & 2.51 & 4.98 & 3685.55 & 2.67 \\
& B4 & 7.28 & 4193.09 & 3.53 & 6.15 & 3906.17 & 3.16 \\
& Ours & 4.05 & 2871.77 & 2.07 & 3.91 & 2872.38 & 2.10 \\
\midrule
\multirow{5}{*}{Sonnet 4.5} & B1 & 1.47 & 49.87 & 0.24 & 1.24 & 50.10 & 0.24 \\
& B2 & 3.44 & 769.87 & 1.29 & 4.62 & 968.85 & 0.91 \\
& B3 & 4.58 & 1797.11 & 2.03 & 4.20 & 1599.78 & 1.46 \\
& B4 & 11.58 & 3327.89 & 3.35 & 8.41 & 2719.29 & 2.80 \\
& Ours & 3.81 & 2037.51 & 1.94 & 4.55 & 2152.45 & 2.09 \\
\bottomrule
\end{tabular}
\caption{\textbf{Computational efficiency by requirement specificity.} Generic requirements provide minimal guidance; specific requirements include domain-specific evaluation criteria.}
\label{tab:efficiency_specificity}
\end{table}

\subsection{Efficiency by Pipeline Stage}
\label{app:efficiency_stages}

Table~\ref{tab:efficiency_stages} decomposes computational costs between planning and implementation phases for two-stage approaches. \textsc{EvalAgent}'s code generation stage is 30--49\% more efficient than B4 (Haiku: 2235K vs 3219K tokens; Sonnet: 1073K vs 2104K tokens), with efficiency gains stemming from evaluation skills that reduce exploration during implementation.

\begin{table}[h]
\centering
\footnotesize
\setlength{\tabcolsep}{4pt}
\renewcommand{\arraystretch}{1.2}
\begin{tabular}{c c c c c c c c}
\toprule
\multirow{2}{*}{LLM} & \multirow{2}{*}{Approach} & \multicolumn{3}{c}{\textbf{Plan Stage}} & \multicolumn{3}{c}{\textbf{Code Stage}} \\
\cmidrule(lr){3-5} \cmidrule(lr){6-8}
& & Time (min) & Tokens (K) & Cost (\$) & Time (min) & Tokens (K) & Cost (\$) \\
\midrule
\multirow{2}{*}{Haiku 4.5} & B4 & 1.55 & 830.50 & 0.88 & 5.17 & 3219.13 & 2.46 \\
& Ours & 0.95 & 636.75 & 0.72 & 3.02 & 2235.32 & 1.37 \\
\midrule
\multirow{2}{*}{Sonnet 4.5} & B4 & 3.63 & 919.36 & 1.06 & 6.36 & 2104.23 & 2.01 \\
& Ours & 1.63 & 1022.05 & 1.12 & 2.55 & 1072.93 & 0.89 \\
\bottomrule
\end{tabular}
\caption{\textbf{Computational efficiency by pipeline stage.} Comparison of planning and code generation phases for two-stage approaches (B4 and \textsc{EvalAgent}).}
\label{tab:efficiency_stages}
\end{table}

\section{Meta-Evaluation Results with Sonnet 4.5 as Meta-Evaluator}
\label{app:sonnet_metaeval}

Table~\ref{tab:win_rate_sonnet_judge} presents full dimension-level win-tie rates when using Sonnet~4.5 as the meta-evaluator, complementing the Opus~4.5 results in Table~\ref{tab:win_rate}. The overall pattern is consistent: \textsc{EvalAgent} achieves strong win-tie rates across all baselines, with the largest advantages in Metric Relevance and Code Quality. Sonnet-as-judge produces slightly more conservative win-tie rates (76--95\% overall) compared to Opus-as-judge (84--97\%), but the ranking of baselines is preserved.

\begin{table*}[htbp]
\centering
\begin{tabular}{c r c c c c c c}
\toprule
& & \multicolumn{6}{c}{\textbf{Win-Tie Rate} (\%)} \\
\graycmidrule{3-8}
LLM & Comparison & \makecell{URF} & \makecell{MR} & \makecell{CQC} & \makecell{PQ} & \makecell{PCA} & \makecell{Overall} \\
\graycmidrulethree{1-1}{2-2}{3-8}
\multirow{6}{*}{\rotatebox{90}{Haiku 4.5}} & Ours \vs B1 & \makecell{81.2\\[-2pt]{\tiny(27W/11T/2L)}} & \makecell{92.5\\[-2pt]{\tiny(37W/0T/3L)}} & \makecell{98.8\\[-2pt]{\tiny(39W/1T/0L)}} & - & - & \makecell{95.0\\[-2pt]{\tiny(38W/0T/2L)}} \\
& \quad \vs B2 & \makecell{80.0\\[-2pt]{\tiny(26W/12T/2L)}} & \makecell{93.8\\[-2pt]{\tiny(37W/1T/2L)}} & \makecell{93.8\\[-2pt]{\tiny(37W/1T/2L)}} & - & - & \makecell{95.0\\[-2pt]{\tiny(38W/0T/2L)}} \\
& \quad \vs B3 & \makecell{63.7\\[-2pt]{\tiny(17W/17T/6L)}} & \makecell{83.8\\[-2pt]{\tiny(33W/1T/6L)}} & \makecell{91.2\\[-2pt]{\tiny(36W/1T/3L)}} & - & - & \makecell{85.0\\[-2pt]{\tiny(34W/0T/6L)}} \\
& \quad \vs B4 & \makecell{51.2\\[-2pt]{\tiny(4W/33T/3L)}} & \makecell{85.0\\[-2pt]{\tiny(32W/4T/4L)}} & \makecell{95.0\\[-2pt]{\tiny(38W/0T/2L)}} & \makecell{85.0\\[-2pt]{\tiny(34W/0T/6L)}} & \makecell{53.8\\[-2pt]{\tiny(18W/7T/15L)}} & \makecell{90.0\\[-2pt]{\tiny(36W/0T/4L)}} \\
\graycmidrule{1-8}
\multirow{6}{*}{\rotatebox{90}{Sonnet 4.5}} & Ours \vs B1 & \makecell{83.8\\[-2pt]{\tiny(29W/9T/2L)}} & \makecell{90.0\\[-2pt]{\tiny(36W/0T/4L)}} & \makecell{92.5\\[-2pt]{\tiny(37W/0T/3L)}} & - & - & \makecell{90.0\\[-2pt]{\tiny(36W/0T/4L)}} \\
& \quad \vs B2 & \makecell{76.2\\[-2pt]{\tiny(27W/7T/6L)}} & \makecell{85.0\\[-2pt]{\tiny(34W/0T/6L)}} & \makecell{92.5\\[-2pt]{\tiny(37W/0T/3L)}} & - & - & \makecell{85.0\\[-2pt]{\tiny(34W/0T/6L)}} \\
& \quad \vs B3 & \makecell{63.7\\[-2pt]{\tiny(18W/15T/7L)}} & \makecell{73.8\\[-2pt]{\tiny(28W/3T/9L)}} & \makecell{76.2\\[-2pt]{\tiny(30W/1T/9L)}} & - & - & \makecell{76.2\\[-2pt]{\tiny(30W/1T/9L)}} \\
& \quad \vs B4 & \makecell{60.0\\[-2pt]{\tiny(13W/22T/5L)}} & \makecell{81.2\\[-2pt]{\tiny(32W/1T/7L)}} & \makecell{87.5\\[-2pt]{\tiny(35W/0T/5L)}} & \makecell{73.8\\[-2pt]{\tiny(28W/3T/9L)}} & \makecell{56.2\\[-2pt]{\tiny(20W/5T/15L)}} & \makecell{85.0\\[-2pt]{\tiny(34W/0T/6L)}} \\
\bottomrule
\end{tabular}
\caption{
\textbf{Meta-evaluation results (Sonnet 4.5 as meta-evaluator).} PQ/PCA not applicable (-) for B1, B2, B3 (no planning). Win-Tie Rate = (wins + 0.5$\times$ties) / total $\times$ 100; higher is better for \textsc{EvalAgent}. Sonnet-as-judge yields slightly more conservative rates than Opus-as-judge, but preserves the same baseline ranking.
}
\label{tab:win_rate_sonnet_judge}
\end{table*}

\section{Meta-Evaluation Consistency Analysis}
\label{app:consistency}

This section provides detailed consistency analysis for the meta-evaluator, examining both cross-model agreement and run-to-run stability.

\subsection{Cross-Model Consistency}
\label{app:cross_model}
Beyond human alignment, we assess meta-evaluator reliability through cross-model consistency: whether different LLM models serving as meta-evaluators reach the same conclusions when evaluating identical evaluation pairs.

We ran the complete meta-evaluation pipeline using two distinct models: Claude Opus 4.5 and Claude Sonnet 4.5. Both models evaluated the same pairwise comparisons across four baseline configurations (LLM-Singleturn (B1), Agent-Sourcecode (B2), Agent-Onestage (B3), and Agent-Twostage (B4) versus \textsc{EvalAgent}).

\begin{table}[h]
\centering
\small
\begin{tabular}{rccc}
\toprule
\multicolumn{4}{c}{\textbf{Win-Tie Rate by Meta-Evaluator LLM}} \\
\graycmidrule{1-4}
Comparison & Opus 4.5 & Sonnet 4.5 & Agreement \\
\graycmidrule{1-4}
Ours \vs B1 & 93.1\% & 91.9\% & 89.4\% \\
\vs B2 & 96.9\% & 89.6\% & 87.1\% \\
\vs B3 & 88.8\% & 80.0\% & 81.3\% \\
\vs B4 & 87.5\% & 86.9\% & 88.8\% \\
\graycmidrule{1-4}
\textbf{Overall} & \textbf{91.6\%} & \textbf{87.1\%} & \textbf{86.6\%} \\
\bottomrule
\end{tabular}
\caption{\textbf{Win-tie rates by meta-evaluator model.} Agreement reports the percentage of individual evaluations on which both meta-evaluators selected the same outcome. Results aggregated across two evaluator LLMs (Haiku 4.5, Sonnet 4.5), two requirement types (generic, specific), and three independent runs.}
\label{tab:cross_model_winrates_app}
\end{table}

As shown in Table~\ref{tab:cross_model_winrates_app}, both models reach qualitatively identical conclusions about relative evaluation quality: \textsc{EvalAgent} achieves win-tie rates above 80\% against every baseline under both meta-evaluators, with inter-model gaps ranging from 0.6 to 8.8 percentage points. The overall cross-model agreement rate of 86.6\% confirms that the two meta-evaluators concur on the vast majority of individual judgments. Notably, the ranking of baselines by difficulty is preserved across models---both identify B1 and B2 as the easiest comparisons and B3 and B4 as the most competitive---indicating that the meta-evaluators capture the same underlying quality gradient.

This cross-model consistency provides complementary evidence to human validation. While human studies establish alignment with expert judgment, cross-model agreement demonstrates that meta-evaluation outcomes are robust to the choice of underlying LLM.

\subsection{Run-to-Run Consistency}
\label{app:run_consistency}
A third signal of meta-evaluator reliability is run-to-run consistency: whether the same meta-evaluator model produces consistent judgments across independent runs. High consistency indicates that evaluations reflect stable reasoning rather than stochastic artifacts of sampling.

We executed the complete meta-evaluation pipeline three times using Claude Opus 4.5 across  all 40 comparison configurations (20 agents × 2 evaluation requirement types).
Across the evaluations, all three runs agreed on 76.3\% of cases and the average pairwise agreement was 84.2\%.

\begin{table}[h]
\centering
\begin{tabular}{lcc}
\toprule
\textbf{Dimension} & \textbf{3-Way Agreement} & \textbf{Pairwise Avg} \\
\midrule
Code Quality \& Complexity & 84.6\% & 89.7\% \\
Metric Relevance & 79.5\% & 86.3\% \\
User Requirement Fulfillment & 75.0\% & 83.3\% \\
\midrule
\textbf{Overall} & \textbf{76.3\%} & \textbf{84.2\%} \\
\bottomrule
\end{tabular}
\caption{\textbf{Per-dimension run-to-run agreement rates} for the Claude Opus 4.5 meta-evaluator across three independent runs and 40 comparison configurations (20 agents × 2 evaluation requirement types). Three-way agreement indicates all three runs selected the same winner; pairwise average reports the mean agreement across all three run pairs.}
\label{tab:run_consistency_dimensions}
\end{table}

Table~\ref{tab:run_consistency_dimensions} presents three-way and pairwise agreement rates stratified by evaluation dimension, averaged across all 40 configurations. Code Quality \& Complexity exhibits the highest consistency (84.6\% three-way, 89.7\% pairwise), likely because code-level properties offer more concrete, less ambiguous evidence for comparison. Metric Relevance and User Requirement Fulfillment show moderately lower but still strong agreement, consistent with the greater interpretive latitude these dimensions afford.

Overall, the results demonstrate that the meta-evaluator produces stable judgments across independent runs: a 76.3\% three-way agreement rate across 40 configurations substantially exceeds the 33.3\% baseline expected from random three-way choices among three outcomes (A wins, B wins, tie). The accompanying 84.2\% pairwise average further confirms that disagreements, when they occur, are typically confined to a single divergent run rather than reflecting systematic instability. These findings support the conclusion that our meta-evaluation methodology yields reproducible judgments suitable for comparing evaluation systems at scale.

\section{Code Examples}
\label{app:code_examples}
This section provides concrete examples supplementing the qualitative analysis in Section~\ref{sec:qualitative}: directory structure and code comparisons (Figures~\ref{fig:dir_comparison}--\ref{fig:code_comparison}), trace-based \vs source-code evaluation, the impact of Context7 on API correctness, and multi-agent evaluation with attribution analysis.

\subsection{Directory Structure and Code Comparison}
\label{app:dir_code_comparison}

\begin{figure*}[htbp]
\centering
\begin{minipage}[c]{0.52\textwidth}
\centering
\begin{forest}
for tree={
  font=\ttfamily\scriptsize,
  grow'=0,
  child anchor=west,
  parent anchor=south,
  anchor=west,
  calign=first,
  edge path={
    \noexpand\path [draw, \forestoption{edge}]
    (!u.south west) +(6pt,0) |- node[fill,inner sep=1pt] {} (.child anchor)\forestoption{edge label};
  },
  before typesetting nodes={
    if n=1
      {insert before={[,phantom]}}
      {}
  },
  fit=band,
  before computing xy={l=10pt},
  s sep=1pt,
  inner xsep=1pt,
  inner ysep=0.5pt,
}
[eval/ {\normalfont\textcolor{gray}{\scriptsize 2{,}290 LOC, 22 files}}
  [agent\_wrapper/
    [\_\_init\_\_.py]
    [city\_agent\_wrapper.py {\normalfont\textcolor{gray}{\scriptsize(133)}}]
    [trace\_analyzer.py]
  ]
  [config/
    [\_\_init\_\_.py]
    [eval\_config.py]
    [metrics\_config.py]
  ]
  [evaluators/
    [\_\_init\_\_.py]
    [accuracy\_evaluator.py]
    [coverage\_evaluator.py]
    [functionality\_evaluator.py]
    [performance\_evaluator.py]
    [quality\_evaluator.py]
    [robustness\_evaluator.py]
  ]
  [generators/
    [\_\_init\_\_.py]
    [\textcolor{red}{ground\_truth\_generator.py} {\normalfont\textcolor{gray}{\scriptsize(130)}}]
    [\textcolor{red}{test\_data\_generator.py} {\normalfont\textcolor{gray}{\scriptsize(151)}}]
  ]
  [utils/
    [\_\_init\_\_.py]
    [logger.py]
    [metrics\_calculator.py]
    [\textcolor{red}{report\_generator.py} {\normalfont\textcolor{gray}{\scriptsize(127)}}]
  ]
  [evaluation\_orchestrator.py]
  [run\_evaluation.py]
]
\end{forest}
\subcaption{B4: \texttt{city\_agent} --- 5 packages, 22 files. {\color{red}Red} = dead code.}
\label{fig:dir_b4}
\end{minipage}%
\hfill
\begin{minipage}[c]{0.42\textwidth}
\centering
\begin{forest}
for tree={
  font=\ttfamily\scriptsize,
  grow'=0,
  child anchor=west,
  parent anchor=south,
  anchor=west,
  calign=first,
  edge path={
    \noexpand\path [draw, \forestoption{edge}]
    (!u.south west) +(6pt,0) |- node[fill,inner sep=1pt] {} (.child anchor)\forestoption{edge label};
  },
  before typesetting nodes={
    if n=1
      {insert before={[,phantom]}}
      {}
  },
  fit=band,
  before computing xy={l=10pt},
  s sep=1pt,
  inner xsep=1pt,
  inner ysep=0.5pt,
}
[eval/ {\normalfont\textcolor{gray}{\scriptsize 207 LOC, 2 files}}
  [metrics.py]
  [run\_evaluation.py]
]
\end{forest}
\subcaption{\textsc{EvalAgent}: \texttt{city\_agent} --- 2 files (19/20 agents).}
\label{fig:dir_ea}
\end{minipage}
\caption{\textbf{Directory structure comparison.} B4 generates a 5-package project (2{,}290~LOC) with dead-code modules unreachable from the entry point. \textsc{EvalAgent} produces 2 files (207~LOC)---an 11$\times$ reduction.}
\label{fig:dir_comparison}
\end{figure*}

\begin{figure*}[htbp]
\centering
\begin{minipage}[t]{0.48\textwidth}
\captionsetup[algorithm]{name=Code}
\begin{algorithm}[H]
\caption{B4: keyword heuristic (\texttt{singleagent\_trip\_crew})}
\label{alg:b4_keyword}
\begin{lstlisting}[language=python,basicstyle=\fontsize{6.5pt}{7.5pt}\ttfamily\bfseries]
class ContentAnalyzer:
  ACTION_VERBS = {
    'visit', 'explore', 'see', 'discover',
    'enjoy', 'experience', 'try', 'taste',
    'sample', 'check out', 'head to',
    'go to', 'walk', 'hike', 'bike',
    'drive', 'take', 'ride', 'dine', 'eat',
    'shop', 'browse', 'relax', 'swim',
    'tour', 'wander', 'stroll', 'climb',
    'watch', 'attend'  # 30 verbs
  }
  LOCATION_KEYWORDS = {
    'temple', 'museum', 'park', 'beach',
    'fort', 'palace', 'market', ...
  }                     # 25 keywords
  PERSONALIZATION_KEYWORDS = {
    'age', 'year old', 'budget',
    'preference', 'interest', ...
  }                     # 10 keywords

  def _count_action_verbs(self) -> int:
    count = 0
    for verb in self.ACTION_VERBS:
      pattern = rf'\b{verb}\b'
      count += len(
        re.findall(pattern, self.output_lower))
    return count
\end{lstlisting}
\end{algorithm}
\end{minipage}%
\hfill
\begin{minipage}[t]{0.48\textwidth}
\captionsetup[algorithm]{name=Code}
\begin{algorithm}[H]
\caption{\textsc{EvalAgent}: LLM-as-Judge (\texttt{singleagent\_trip\_crew})}
\label{alg:ea_geval}
\begin{lstlisting}[language=python,basicstyle=\fontsize{6.5pt}{7.5pt}\ttfamily\bfseries]
class ItineraryCompletenessMetric(BaseMetric):
  def __init__(self, threshold=0.7):
    self.threshold = threshold
    self.geval = GEval(
      name="Itinerary Completeness",
      criteria="""Evaluate whether
the travel plan includes:
1. Destination research (attractions,
   events, transport, weather, safety)
2. Local recommendations (hidden gems,
   customs, timing tips)
3. Restaurant recommendations
4. Day-by-day itinerary with timing
   and logistics

Score 1.0 if all components present
and comprehensive, 0.0 if missing""",
      evaluation_params=[
        LLMTestCaseParams.INPUT,
        LLMTestCaseParams.ACTUAL_OUTPUT],
      model="us.anthropic.claude-sonnet-4-..."
    )

  def measure(self, test_case):
    self.geval.measure(test_case)
    self.score = self.geval.score
    self.reason = self.geval.reason
    return self.score
\end{lstlisting}
\end{algorithm}
\end{minipage}
\caption{\textbf{Keyword heuristic \vs LLM-as-Judge.} B4 counts regex matches against hardcoded word lists; \textsc{EvalAgent} uses GEval with structured criteria. Both from \texttt{singleagent\_trip\_crew}.}
\label{fig:code_comparison}
\end{figure*}

\subsection{Trace-Based vs Source-Code Evaluation}
\label{app:trace_vs_source}

\paragraph{Source-code limitation.}
Without execution traces, Agent-Sourcecode (B2) must create synthetic test data from static analysis. Code~\ref{alg:source_code_synthetic} shows this limitation, generating hypothetical test cases without observing actual agent behavior.
\begin{figure}[htbp]
\begin{minipage}[t]{\textwidth}
\captionsetup[algorithm]{name=Code}
\begin{algorithm}[H]
   \caption{Source-code approach (B2) - Synthetic test data generation}
   \label{alg:source_code_synthetic}
    \begin{lstlisting}[language=python]
test_task = {
    "task_id": "test_001",
    "test_cases": [
        {"input": {"n": 0}, "expected": 1},
        {"input": {"n": 5}, "expected": 120}
    ]
}
    \end{lstlisting}
\end{algorithm}
\end{minipage}
\end{figure}
The synthetic data assumes factorial computation based on function signatures alone. If the agent handles edge cases differently (\eg, returning \texttt{None} for negative inputs), this evaluation misses such behaviors.

\paragraph{Trace benefits.}
The Agent-Onestage (B3) parses actual OTEL traces from genuine agent runs. Code~\ref{alg:trace_extraction_code} demonstrates trace extraction for the medical document processor, capturing real prompts, tool calls, and outputs.
\begin{figure}[htbp]
\begin{minipage}[t]{\textwidth}
\captionsetup[algorithm]{name=Code}
\begin{algorithm}[H]
   \caption{Trace extraction code for medical document processor agent}
   \label{alg:trace_extraction_code}
    \begin{lstlisting}[language=python]
def extract_from_trace(trace_path: str) -> TraceData:
    for span in scope_span.get("spans", []):
        for event in span.get("events", []):
            # Extract user input from trace
            if event.get("name") == "gen_ai.user.message":
                input_text = extract_content(event)

            # Extract tool calls and their outputs
            if span.get("name") == "execute_tool link_icd":
                # Extract ICD-10 diagnoses with codes
                for entity in parse_tool_output(event):
                    diagnoses.append(ExtractedEntity(
                        entity_type="diagnosis",
                        value=entity.get("diagnosis", ""),
                        code=entity.get("ICD10_code", ""),
                        confidence_score=entity.get("confidence_score")
                    ))
            elif span.get("name") == "execute_tool link_rx":
                # Extract medications with RxNorm codes
                medications.append(ExtractedEntity(...))
    \end{lstlisting}
\end{algorithm}
\end{minipage}
\end{figure}
Code~\ref{alg:trace_extraction_code} captures user input from \texttt{gen\_ai.user.message} events, real tool invocations (\texttt{link\_icd}, \texttt{link\_rx}, \texttt{link\_snomed}), extracted entities with codes (\eg, \texttt{\{"diagnosis": "Type 2 Diabetes", "ICD10\_code": "E11.9"\}}), and the final response. This enables the metrics in Code~\ref{alg:trace_metrics_code}, which validates actual extracted entities against ground truth.
\begin{figure}[htbp]
\begin{minipage}[t]{\textwidth}
\captionsetup[algorithm]{name=Code}
\begin{algorithm}[H]
   \caption{Metrics only possible with actual trace data}
   \label{alg:trace_metrics_code}
    \begin{lstlisting}[language=python]
def evaluate_trace(trace_data: TraceData, ground_truth: Dict):
    return {
        "entity_extraction": {
            # Precision/recall on extracted entities
            "diagnoses": calculate_entity_accuracy(
                trace_data.diagnoses, ground_truth["diagnoses"]),
            "medications": calculate_entity_accuracy(
                trace_data.medications, ground_truth["medications"])
        },
        "code_correctness": {
            # Validate medical codes against standards
            "diagnoses": calculate_code_correctness(
                trace_data.diagnoses, ground_truth["diagnoses"])
        }
    }
    \end{lstlisting}
\end{algorithm}
\end{minipage}
\end{figure}
Lines 4--8 compute precision/recall on \emph{actual extracted diagnoses}, validating ``Type 2 Diabetes'' with ICD-10 code ``E11.9'', not merely keyword presence. This granularity requires trace data.

Without traces, B2 relies on heuristics. Code~\ref{alg:source_code_hypothesized} counts keyword occurrences rather than validating semantic correctness.
\begin{figure}[htbp]
\begin{minipage}[t]{\textwidth}
\captionsetup[algorithm]{name=Code}
\begin{algorithm}[H]
   \caption{Source-code approach: hypothesized behavior evaluation}
   \label{alg:source_code_hypothesized}
    \begin{lstlisting}[language=python]
def evaluate_medical_terminology_usage(output: str):
    # Static list - doesn't know what agent actually extracted
    medical_terms = ['diagnosis', 'medication', 'ICD', 'RxNorm']
    found_terms = [t for t in medical_terms if t.lower() in output.lower()]
    return {'medical_terms_found': len(found_terms)}
    \end{lstlisting}
\end{algorithm}
\end{minipage}
\end{figure}
The heuristic cannot distinguish ``The diagnosis is incorrect'' from ``The diagnosis is Type 2 Diabetes''; both score equally. Source-code approaches analyze implementation structure but cannot observe runtime behavior: tool calls, error handling, or execution paths.

\subsection{With vs. Without Context7: Dynamic Documentation Retrieval}
\label{app:context7_examples}
The root cause of failures without Context7 is API drift: library APIs evolve faster than model training cycles. Context7 fetches current documentation at generation time.

\paragraph{Case study: Career Assistant model integration.}
The \texttt{career\_assist} agent requires LLM-as-judge metrics. Code~\ref{alg:context7_without_career} shows a common failure: missing the \texttt{bedrock/} provider prefix without Context7.
\begin{figure}[htbp]
\begin{minipage}[t]{\textwidth}
\captionsetup[algorithm]{name=Code}
\begin{algorithm}[H]
   \caption{Without Context7 - Wrong model identifier format}
   \label{alg:context7_without_career}
    \begin{lstlisting}[language=python]
from deepeval.metrics import BaseMetric, GEval
from deepeval.test_case import LLMTestCase, LLMTestCaseParams

class PositionSkillAlignmentMetric(BaseMetric):
    def __init__(self, threshold: float = 0.7):
        self.threshold = threshold
        self.geval = GEval(
            name="Position-Skill Alignment",
            criteria="Evaluate how well the suggested positions align...",
            evaluation_params=[
                LLMTestCaseParams.INPUT,
                LLMTestCaseParams.ACTUAL_OUTPUT
            ],
            # WRONG: Missing provider prefix 'bedrock/'
            model="us.anthropic.claude-sonnet-4-20250514-v1:0"
        )
    \end{lstlisting}
\end{algorithm}
\end{minipage}
\end{figure}
The model identifier on line 12 requires the \texttt{bedrock/} prefix; this error accounts for most execution failures in Table~\ref{tab:ablation_context7_exec}. Code~\ref{alg:context7_with_career} shows the correct pattern from DeepEval documentation via Context7.
\begin{figure}[htbp]
\begin{minipage}[t]{\textwidth}
\captionsetup[algorithm]{name=Code}
\begin{algorithm}[H]
   \caption{With Context7 - Proper DeepEvalBaseLLM wrapper}
   \label{alg:context7_with_career}
    \begin{lstlisting}[language=python]
from deepeval.metrics import BaseMetric
from deepeval.models import DeepEvalBaseLLM
from litellm import completion

class LiteLLMWrapper(DeepEvalBaseLLM):
    """Custom model wrapper per DeepEval documentation."""
    def __init__(self, model: str = "bedrock/us.anthropic.claude-sonnet-4-20250514-v1:0"):
        self.model = model

    def load_model(self):
        return self.model

    def generate(self, prompt: str) -> str:
        response = completion(
            model=self.model,
            messages=[{"role": "user", "content": prompt}],
            temperature=0.0
        )
        return response.choices[0].message.content

    async def a_generate(self, prompt: str) -> str:
        return self.generate(prompt)

    def get_model_name(self) -> str:
        return self.model

class PositionSkillAlignmentMetric(BaseMetric):
    def __init__(self, threshold: float = 0.6,
                 model: str = "bedrock/us.anthropic.claude-sonnet-4-20250514-v1:0"):
        self.threshold = threshold
        self.model = LiteLLMWrapper(model)  # Proper wrapper instantiation
    \end{lstlisting}
\end{algorithm}
\end{minipage}
\end{figure}
Context7 retrieves DeepEval documentation describing the \texttt{DeepEvalBaseLLM} abstract class (line 2), required method signatures (\texttt{load\_model()}, \texttt{generate()}, \texttt{a\_generate()}, \texttt{get\_model\_name()}), and the correct \texttt{bedrock/} prefix format (line 7).

\paragraph{Case study: Hilton Search location matching.}
The \texttt{hilton\_search\_agent} requires evaluation of location matching accuracy. This case demonstrates how Context7 enables sophisticated LLM-as-judge evaluation versus primitive heuristic fallbacks.

\subparagraph{Without Context7: brittle keyword matching.}
\begin{figure}[htbp]
\begin{minipage}[t]{\textwidth}
\captionsetup[algorithm]{name=Code}
\begin{algorithm}[H]
   \caption{Without Context7 - Brittle keyword matching}
   \label{alg:context7_without_hilton}
    \begin{lstlisting}[language=python]
# Falls back to keyword matching - no LLM evaluation capability
def evaluate_location_matching(query: str, response: str) -> float:
    query_lower = query.lower()
    response_lower = response.lower()

    # Brittle string matching - misses semantic understanding
    if 'san francisco' in query_lower and 'financial district' in query_lower:
        if 'financial district' in response_lower:
            return 1.0
        elif 'san francisco' in response_lower:
            return 0.5

    return 0.0  # Cannot evaluate novel locations
    \end{lstlisting}
\end{algorithm}
\end{minipage}
\end{figure}
The heuristic in Code~\ref{alg:context7_without_hilton} has three critical flaws: hard-coded location checks (lines 5--9) fail for unlisted locations, semantic equivalences like ``Downtown SF'' $\approx$ ``Financial District'' score 0.0, and no reasoning is provided for scores.

\subparagraph{With Context7: LLM-as-Judge evaluation.}
Code~\ref{alg:context7_with_hilton} demonstrates proper LLM-as-judge evaluation using the LiteLLM \texttt{completion()} API documented via Context7.
\begin{figure}[htbp]
\begin{minipage}[t]{\textwidth}
\captionsetup[algorithm]{name=Code}
\begin{algorithm}[H]
   \caption{With Context7 - Proper LLM-as-judge evaluation}
   \label{alg:context7_with_hilton}
    \begin{lstlisting}[language=python]
from deepeval.metrics import BaseMetric
from litellm import completion

class LocationMatchingAccuracy(BaseMetric):
    def __init__(self, threshold=0.7,
                 model="bedrock/us.anthropic.claude-sonnet-4-20250514-v1:0"):
        self.threshold = threshold
        self.model = model

    def measure(self, test_case: LLMTestCase) -> float:
        prompt = f"""Evaluate whether hotel results match location requirements.
User Query: {test_case.input}
Agent Response: {test_case.actual_output}
Consider: City/district match, proximity, general area.
Score 0-1.
Respond JSON: {{"score": <float>, "reason": "<explanation>"}}"""

        response = completion(
            model=self.model,
            messages=[{"role": "user", "content": prompt}],
            temperature=0.0
        )
        result = json.loads(response.choices[0].message.content)
        self.score = result["score"]
        self.reason = result.get("reason", "")
        return self.score
    \end{lstlisting}
\end{algorithm}
\end{minipage}
\end{figure}
The LLM-as-judge approach in Code~\ref{alg:context7_with_hilton} generalizes to any location, captures semantic similarity, and provides reasoning via \texttt{result["reason"]}.

\subsection{Multi-Agent System Evaluation}
\label{app:multiagent_examples}
Multi-agent systems require evaluation of agent attribution, coordination quality, and tool routing.

\paragraph{Agent attribution via OTEL spans.}
\textsc{EvalAgent} uses the \texttt{gen\_ai.agent.name} span attribute to trace contributions across cooperating agents. Code~\ref{alg:warehouse_attribution} demonstrates this pattern for the data warehouse optimizer's three-agent workflow: \texttt{analyzer\_agent} $\rightarrow$ \texttt{rewriter\_agent} $\rightarrow$ \texttt{validator\_agent}.
\begin{figure}[htbp]
\begin{minipage}[t]{\textwidth}
\captionsetup[algorithm]{name=Code}
\begin{algorithm}[H]
   \caption{\textsc{EvalAgent} agent attribution for data\_warehouse\_optimizer}
   \label{alg:warehouse_attribution}
    \begin{lstlisting}[language=python]
def extract_agent_name(span: Dict[str, Any]) -> Optional[str]:
    """Extract agent name from span attributes."""
    for attr in attrs:
        if attr.get('key') == 'gen_ai.agent.name':
            return attr.get('value', {}).get('stringValue')
    return None

class OptimizationQualityMetric:
    """Evaluate quality of optimization suggestions."""
    @staticmethod
    def calculate(result: OptimizationResult) -> Dict[str, Any]:
        details = {'has_index_recommendation': False,
                   'has_query_rewrite': False}
        for suggestion in result.optimization_suggestions:
            if 'index' in sug_type.lower():
                details['has_index_recommendation'] = True
            if 'query_optimization' in sug_type.lower():
                details['has_query_rewrite'] = True
    \end{lstlisting}
\end{algorithm}
\end{minipage}
\end{figure}
Lines 2--6 extract agent names from OTEL span attributes; lines 9--16 evaluate suggestion quality by checking for index recommendations and query rewrites. The \texttt{ToolUsageMetric} (not shown) validates that expected tools are invoked by the correct agents.

\end{document}